\pdfoutput=1
\documentclass[10pt, a4paper, twocolumn, showabstract]{naverlabseurope}

\usepackage{amsmath}
\usepackage{xcolor}
\usepackage{tikz}
\usepackage{units}
\usepackage{makecell}
\newcommand{\ours}[0]{MASt3R-SfM}

\newcommand{\R}{\mathbb{R}} %
\newcommand{\mypar}[1]{\paragraph{#1}}

\newcommand{\duster}[0]{DUSt3R}
\newcommand{\master}[0]{MASt3R}
\newcommand{\I}[1]{I^{#1}} %
\newcommand{\EF}[1]{F^{#1}} %
\newcommand{\K}[1]{K_{#1}} %
\newcommand{\cam}[1]{P_{#1}} %
\newcommand{\D}[1]{D^{#1}} %
\newcommand{\X}[2]{X^{#1,#2}} %
\newcommand{\depth}[1]{Z^{#1}} %

\newcommand{\enc}{\text{Enc}} 
\newcommand{\dec}{\text{Dec}}

\newcommand{\M}{\mathcal{M}} %
\newcommand{\G}{\mathcal{G}} %
\newcommand{\V}{\mathcal{V}} %
\newcommand{\E}{\mathcal{E}} %
\newcommand{\T}{\mathcal{T}} %
\newcommand{\TE}{\mathcal{D}} %

\newcommand{\cX}[1]{\tilde{X}^{#1}} %
\newcommand{\cdepth}[1]{\tilde{Z}^{#1}} %
\newcommand{\cK}[1]{\tilde{K}^{#1}} %

\DeclareMathOperator*{\argmin}{arg\,min}
\newcommand{\eqdef}{\stackrel{\text{def}}{=}}

\newcommand{\bgcolor}[2]{\setlength{\fboxsep}{0pt}\colorbox{#1}{\strut #2}}

\definecolor{cbad}{HTML}{EAC2C2}
\definecolor{cmeh}{HTML}{FFE1C9}
\definecolor{cok}{HTML}{FDF3D0}
\definecolor{cgood}{HTML}{B3D09F}

\pgfdeclarehorizontalshading{rankgrad}{100bp}{
    color(0bp)=(cbad);
    color(20bp)=(cbad);
    color(40bp)=(cmeh);
    color(60bp)=(cok);
    color(80bp)=(cgood);
    color(100bp)=(cgood)
}

\newcommand{\coloredCell}[3]{\definecolor{mycolor}{HTML}{#2}\tikz[baseline=(char.base)]{\node[fill=mycolor,inner ysep=2pt, inner xsep=5pt, minimum width=#1, text width=#1, align=right] (char) {#3};}}

\newcommand{\cellCD}[3][24pt]{\coloredCell{#1}{#2}{#3}}
\newcommand{\cellCDS}[3][24pt]{\cellcolor[HTML]{#2}{\parbox[t]{#1}{\centering #3}}}

\newcommand{\BGcolor}[3][HTML]{\definecolor{mycolor}{HTML}{#2}\bgcolor{mycolor}{#3}}

\usepackage{graphicx}
\usepackage{rotating}
\usepackage{multirow}
\usepackage{multicol}
\usepackage{hyperref}
\usepackage{cleveref}

\makeatletter
\renewcommand{\paragraph}{%
  \@startsection{paragraph}{4}%
  {\z@}{0.3ex \@plus 1ex \@minus .2ex}{-1em}%
  {\normalfont\normalsize\bfseries}%
}

\usepackage{xspace}
\DeclareRobustCommand\onedot{\futurelet\@let@token\@onedot}
\def\@onedot{\ifx\@let@token.\else.\null\fi\xspace}
\def\eg{\emph{e.g}\onedot} 
\def\ie{\emph{i.e}\onedot} 
 
\def\etc{\emph{etc}\onedot} 
\def\wrt{w.r.t\onedot}

\makeatother

\expandafter\def\expandafter\normalsize\expandafter{%
    \normalsize%
    \setlength\abovedisplayskip{3pt}%
    \setlength\belowdisplayskip{3pt}%
    \setlength\abovedisplayshortskip{3pt}%
    \setlength\belowdisplayshortskip{3pt}%
}

\title{\ours{}: a Fully-Integrated Solution for Unconstrained Structure-from-Motion}

\correspondingauthor{[vincent.leroy,jerome.revaud]@naverlabs.com}

\authors{Bardienus DUISTERHOF, \authsep Lojze ZUST, \authsep Philippe WEINZAEPFEL, \authsep Vincent Leroy, \authsep Yohann Cabon, and \authsep Jerome Revaud}
\affiliations{NAVER LABS Europe}
\contributions{} %
\website{https://github.com/naver/mast3r}
\websiteref{\href{https://github.com/naver/mast3r}}

\begin{abstract}

Structure-from-Motion (SfM), a task aiming at jointly recovering camera poses and 3D geometry of a scene given a set of images, remains a hard problem with still many open challenges despite decades of significant progress. The traditional solution for SfM consists of a complex pipeline of minimal solvers which tends to propagate errors and fails when images do not sufficiently overlap, have too little motion, etc. 
Recent methods have attempted to revisit this paradigm, but we empirically show that they fall short of fixing these core issues. In this paper, we propose instead to build upon a recently released foundation model for 3D vision that can robustly produce local 3D reconstructions and accurate matches. We introduce a low-memory approach to accurately align these local reconstructions in a global coordinate system. We further show that such foundation models can serve as efficient image retrievers without any overhead, reducing the overall complexity from quadratic to linear. Overall, our novel SfM pipeline is simple, scalable, fast and truly unconstrained, i.e. it can handle any collection of images, ordered or not. Extensive experiments on multiple benchmarks show that our method provides steady performance across diverse settings, especially outperforming existing methods in small- and medium-scale settings.

\end{abstract}

\begin{document}
\maketitle

\section{Introduction}
\label{sec:intro}

\begin{figure}[t]
    \centering
    
    \includegraphics[width=\linewidth]{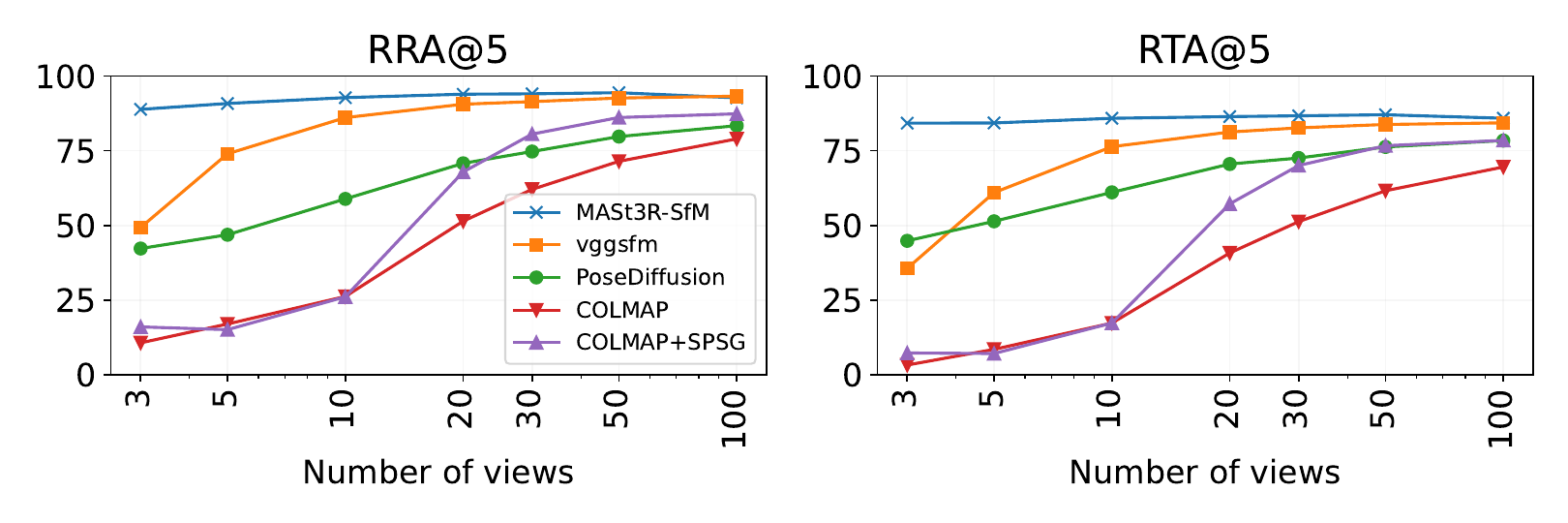}
    \resizebox{\linewidth}{!}{
        \includegraphics[height=0.2\linewidth]{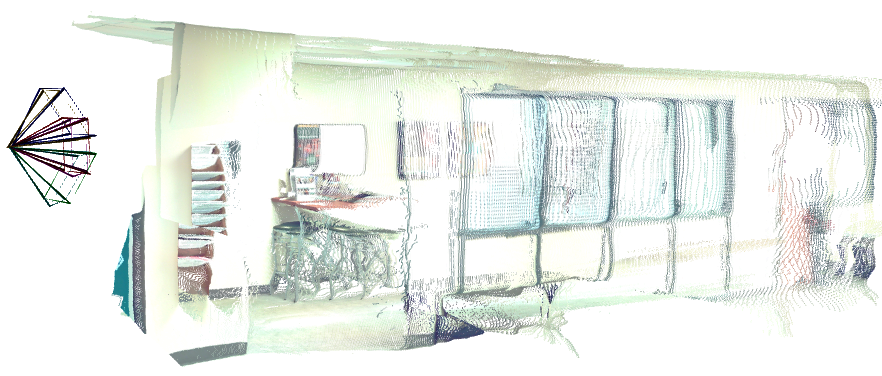}
        \includegraphics[height=0.2\linewidth]{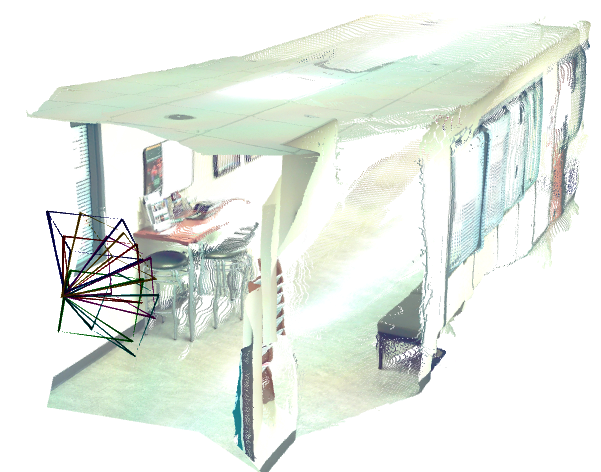}
    }
    \vspace{-0.6cm}
    \caption{
        \textbf{Top}: Relative rotation (RRA) and translation (RTA) accuracies on the CO3Dv2 dataset when varying the number of input views with random subsampling (the more views, the larger they overlap). 
        In contrast to our competitors, \ours{} offers nearly constant performance on the full range, even for very few views.
        \textbf{Bottom}: \ours{} also works \emph{without motion}, \ie in purely rotational settings.
        We show here a reconstruction from 6 views sharing the same optical center.
    }
    \label{fig:teaser}
\end{figure}

Structure-from-Motion (SfM) is a long-standing problem of computer vision that aims to estimate the 3D geometry of a scene as well as the parameters of the cameras observing it, given the images from each camera~\cite{hartley}. 
Since it conveniently provides jointly for cameras and map, it constitutes an essential component for many practical computer vision applications, such as navigation (mapping and visual localization~\cite{colmapsfm,orbslam,orbslam3}), dense multi-view stereo reconstruction (MVS)~\cite{colmapmvs,yang2020cost,wang2021patchmatchnet,peng2022rethinking}, novel view synthesis~\cite{nerf,mipnerf360,3dgs}, auto-calibration~\cite{selfcalib} or even archaeology~\cite{peppa2018archaeological,thrun2002prob}. 

In reality, SfM is a ``needle in a haystack" type of problem, typically involving a highly non-convex objective function with many local minima~\cite{goos_bundle_2000}. 
Since finding the global minimum in such a landscape is too challenging to be done directly, traditional SfM approaches such as COLMAP~
\cite{colmapsfm} have been decomposing the problem as a series (or \emph{pipeline}) of minimal problems, \eg keypoint extraction and matching, relative pose estimation, and incremental reconstruction with triangulation and bundle adjustment.
The presence of outliers, such as wrong pixel matches, poses additional challenges and compels existing methods to repeatedly resort to hypothesis formulation and verification at multiple occasions in the pipeline, typically with RAndom SAmple Consensus %
(RANSAC) or its many flavors~\cite{ransac,loransac,MAGSACpp,mlesac,GC-RANSAC,GD-RANSAC}.
This approach has been the standard for several decades, yet it remains brittle and fails when the input images do not sufficiently overlap, or when motion (\ie translation) between viewpoints is insufficient~\cite{SfMfailureNoMotion,SfMfailureLowOverlap}.

Recently, a set of innovative methods propose to revisit SfM in order to alleviate the heavy complexity of the traditional pipeline and solve its shortcomings.
VGGSfM~\cite{vggsfm}, for instance, introduces an end-to-end differentiable version of the pipeline, simplifying some of its components. %
Likewise, detector-free SfM~\cite{detfreeSfm} replaces the keypoint extraction and matching step of the classical pipeline with learned components.
These changes must, however, be put into perspective, as they do not fundamentally challenge the overall structure of the traditional pipeline.
In comparison, FlowMap~\cite{flowmap} and Ace-Zero~\cite{acezero} independently propose a radically novel type of approach to solve SfM, which is based on simple first-order gradient descent of a global loss function.
Their trick is to train a geometry regressor network during scene optimization as a way to reparameterize and regularize the scene geometry.
Unfortunately, this type of approach only works in certain configurations, namely for input images exhibiting high overlap and low illumination variations.
Lastly, \duster{}~\cite{dust3r,mast3r} demonstrates that a single forward pass of a transformer architecture can provide for a good estimate of the geometry and cameras parameters of a small two-image scene. 
These particularly robust estimates can then be stitched together again using simple gradient descent, allowing to relax many of the constraints mentioned earlier.
However it yields rather imprecise global SfM reconstructions and does not scale well.

In this work, we propose \ours{}, a fully-integrated SfM pipeline that can handle completely unconstrained input image collections, \ie ranging from a single view to large-scale scenes, possibly without any camera motion as illustrated in \cref{fig:teaser}.
We build upon the recently released \duster{}~\cite{dust3r}, a foundation model for 3D vision, and more particularly on its recent extension \master{} that is able to perform local 3D reconstruction and matching in a single forward pass~\cite{mast3r}.
Since \master{} is fundamentally limited to processing image pairs, it scales poorly to large image collections.
To remedy this, we hijack its frozen encoder to perform fast image retrieval with negligible computational overhead, resulting in a scalable SfM method with quasi-linear complexity in the number of images.
Thanks to the robustness of \master{} to outliers, the proposed method is able to completely get rid of RANSAC.
The SfM optimization is carried out in two successive gradient descents based on frozen local reconstructions output by \master{}: 
first, using a matching loss in 3D space; then with a 2D reprojection loss to refine the previous estimate. 
Interestingly, our method goes beyond structure-from-motion, as it works even when there is \emph{no motion} (\ie purely rotational case), as illustrated in \cref{fig:teaser}.

In summary, we make three main contributions.
First, we propose \ours{}, a full-fledged SfM pipeline able to process unconstrained image collections. 
To achieve linear complexity in the number of images, we show as second contribution how the encoder from \master{} can be exploited for large-scale image retrieval.
Note that our entire SfM pipeline is training-free, provided an off-the-shelf \master{} checkpoint.
Lastly, we conduct an extensive benchmarking on a diverse set of datasets, showing that existing approaches are still prone to failure in small-scale settings, despite significant progress.
In comparison, \ours{} demonstrates state-of-the-art performance in a wide range of conditions, as illustrated in \cref{fig:teaser}.

\section{Related Works}
\label{sec:related}

\mypar{Traditional SfM.}
At the core of Structure-from-Motion (SfM) lies matching and Bundle Adjustment (BA). 
Matching, \ie the task of finding pixel correspondences across different images observing the same 3D points, has been extensively studied in the past decades, beginning from handcrafted keypoints~\cite{sift,surf,orb} and more recently being surpassed by data-driven strategies~\cite{r2d2,superpoint,superglue,cotr,loftr,MatchFormer,aspanformer,dkm,roma}.
Matching is critical for SfM, since it builds the basis to formulate a loss function to minimize during BA.
BA itself aims at minimizing reprojection errors for the correspondences extracted during the matching phase by jointly optimizing the positions of 3D point and camera parameters. 
It is usually expressed as a non-linear least squares problem~\cite{ceres}, known to be brittle in the presence of outliers and prone to fall into suboptimal local minima if not provided with a good initialization~\cite{phototourism,BAinTheLarge}.
For all these reasons, traditional SfM pipelines like COLMAP are heavily handcrafted in practice~\cite{colmapsfm,pixsfm,detfreeSfm}. 
By triangulating 3D points to provide an initial estimate for BA, they incrementally build a scene, adding images one by one by formulating hypothesis and discarding the ones that are not verified by the current scene state. 
Due to the large number of outliers, and the fact that the structure of the pipeline tends to propagate errors rather than fix them, robust estimators like RANSAC are extensively used for relative pose estimation, keypoint track construction and multi-view triangulation~\cite{colmapsfm}.

\mypar{SfM revisited.} 
There has been a recent surge of methods aiming to simplify or even completely revisit the traditional SfM pipeline~\cite{dust3r,vggsfm,acezero,flowmap,detfreeSfm}.
The recently proposed FlowMap and Ace-Zero, for instance, both rely on the idea of training a regressor network at test time. %
In the case of FlowMap~\cite{flowmap}, this network predicts depthmaps, while for Ace-Zero~\cite{acezero} it regresses dense 3D scene coordinates.
While this type of approach is appealing, it raises several problems such as scaling poorly and depending on many off-the-shelf components for FlowMap. %
Most importantly, both methods only apply to constrained settings where the input image collections offers enough uniformity and continuity in terms of viewpoints and illuminations.
This is because the regressor network is only able to propagate information incrementally from one image to other tightly similar images.
As a result, they cannot process unordered image collections with large viewpoint and illumination disparities.
On the other hand, VGGSfM, Detector-Free SfM (DF-SfM) and \duster{} cast the SfM problem in a more traditional manner by relying on trained neural components that are kept frozen at optimization time.
VGGSfM~\cite{vggsfm}, for its part, essentially manages to train end-to-end all components of the traditional SfM pipeline but still piggybacks itself onto handcrafted solvers for initializing keypoints, cameras and to triangulate 3D points.
As a result, it suffers from the same fundamental issues than traditional SfM, \eg it struggles when there are few views or little camera motion.
Likewise, DF-SfM~\cite{detfreeSfm} improves for texture-less scenes thanks to relying on trainable dense pairwise matchers, but sticks to the overall COLMAP pipeline.
Finally, \duster{}~\cite{dust3r} is a foundation model for 3D vision that essentially decomposes SfM into two steps: local reconstruction for every image pair in the form of pointmaps, and global alignment of all pointmaps in world coordinates.
While the optimization appears considerably simpler than for previous approaches (\ie not relying on external modules, and carried out by minimizing a global loss with first-order gradient descent), it unfortunately yields rather imprecise estimates and does not scale well.
Its recent extension \master{}~\cite{mast3r} adds pixel matching capabilities and improved pointmap regression, but does not address the SfM problem.
In this work, we fill this gap and present a fully-integrated SfM pipeline based on \master{} that is both precise and scalable. %

\mypar{Image Retrieval for SfM.}
Since matching is essentially considering pairs in traditional SfM, it has a quadratic complexity which becomes prohibitive for large image collections.
Several SfM approaches have proposed to leverage faster, although less precise, image comparison techniques relying on comparing global image descriptors, \eg AP-GeM~\cite{apgem} for Kapture~\cite{kapture} or by distilling NetVLAD~\cite{netvlad} for HLoc~\cite{hloc}.
The idea is to cascade image matching in two steps: first, a coarse but fast comparison is carried out between all pairs (usually by computing the similarity between global image descriptors), and for image pairs that are similar enough, a second stage of costly keypoint matching is then carried out.
This is arguably much faster and scalable.
In this paper, we adopt the same strategy, but instead of relying on an external off-the-shelf module, we show that we can simply exploit the frozen \master{}'s encoder for this purpose, considering the token features as local features and directly performing efficient retrieval with Aggregated Selective Match Kernels (ASMK)~\cite{asmk}.

\begin{figure*}
    \centering
    \includegraphics[width=1\linewidth,trim=0 414 0 0, clip]{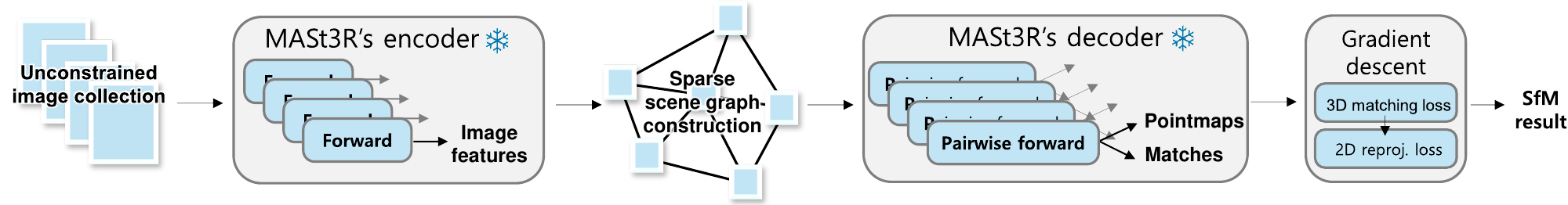} \\[-0.2cm]
    \caption{\textbf{Overview of the proposed \ours{} method}.
    Given an unconstrained image collections, possibly small (1 image) or large ($>1000$ images), we start by computing a sparse scene graph using efficient image retrieval techniques given a frozen \master{}'s per-image features.
    We then compute local 3D reconstruction and matches for each edge using again a frozen \master{}'s decoder.
    Global optimization proceeds with gradient descent of a matching loss in 3D space, followed by refinement in terms of 2D reprojection error.
    }
    \vspace{-0.2cm}
    \label{fig:overview}
\end{figure*}

\section{Preliminaries}
\label{sec:prelim}

The proposed method builds on the recently introduced \master{} model which, given two input images $\I{n},\I{m}\in \R^{H \times W \times 3}$, performs joint \emph{local 3D reconstruction} and \emph{pixel-wise matching}~\cite{mast3r}.
We assume here for simplicity that all images have the same pixel resolution $W\times H$, but of course they can differ in practice. 
In the next section, we show how to leverage this powerful \emph{local} predictor for achieving large-scale \emph{global} 3D reconstruction.

At a high level, \master{} can be viewed as a function $f(\I{n},\I{m}) \equiv \dec(\enc(\I{n}),\enc(\I{m}))$, where $\enc(I)\rightarrow \EF{}$ denotes the Siamese ViT encoder that represents image $\I{}$ as a feature map of dimension $d$, width $w$ and height $h$, $\EF{}\in\R^{h \times w \times d}$, and $\dec(\EF{n},\EF{m})$ denotes twin ViT decoders that %
regresses pixel-wise pointmaps $X$ and local features $\D{}$ for each image, as well as their respective corresponding confidence maps.
These outputs intrinsically contain rich geometric information from the scene, to the extent that camera intrinsics %
and (metric) depthmaps %
can straightforwardly be recovered from the pointmap, see \cite{dust3r} for details.
Likewise, we can recover sparse correspondences (or \emph{matches}) by application of the fastNN algorithm described in~\cite{mast3r} with the regressed local feature maps $D^n, D^m$.
More specifically, the fast NN searches for a subset of reciprocal correspondences from two feature maps $\D{n}$ and $\D{m}$ by initializing seeds on a regular pixel grid %
and iteratively converging to mutual correspondences.
We denote these correspondences between $I^{n}$ and $I^{m}$ as $\mathcal{M}^{n,m}=\{y_{c}^{n}\leftrightarrow y_{c}^{m}\}_{c=1..|\mathcal{M}^{n,m}|}$, where $y_c^n,y_c^m\in \mathbb{N}^2$ denotes a pair of matching pixels.

\section{Proposed Method}
\label{sec:method}

Given an unordered collection of $N$ images $\V=\{I^{n}\}_{1\le n\le N}$ of a static 3D scene, captured with respective cameras $\mathcal{K}_{n}=(\K{n},\cam{n})$, 
where $\K{n}\in R^{3\times3}$ denotes the intrinsic parameters (\ie calibration in term of focal length and principal point) and $\cam{n}\in R^{4\times4}$ its world-to-camera pose,
our goal is to recover all cameras parameters $\{\mathcal{K}_{n}\}$ as well as the underlying 3D scene geometry $\{X^{n}\}$, with $X^{n}\in\mathbb{R}^{W\times H\times3}$ a pointmap relating each pixel $y=(i,j)\in\mathbb{N}^{2}$ from $\I{n}$ to its corresponding 3D point $X_{i,j}^{n}$ in the scene expressed in a world coordinate system. 

\mypar{Overview.}

We present a novel large-scale 3D reconstruction approach consisting of four steps outlined in \cref{fig:overview}.
First, we construct a co-visibility graph using efficient and scalable image retrieval techniques.
Edges of this graph connect pairs of likely-overlapping images.
Second, we perform pairwise local 3D reconstruction and matching using \master{} for each edge of this graph.
Third, we coarsely align every local pointmap in the same world coordinate system using gradient descent with a matching loss in 3D space.
This serves as initialization for the fourth step, wherein we perform a second stage of global optimization, this time minimizing 2D pixel reprojection errors.
We detail each step below.

\subsection{Scene graph}
\label{sub:scenegraph}

We first aim at spatially relating scene objects seen under different viewpoints.
Traditional SfM methods use efficient and scalable keypoint matching for that purpose, thereby building point tracks spanning multiple images.
However, \master{} is originally a pairwise image matcher, which has quadratic complexity in the number $N$ of images and therefore becomes infeasible for large collections if done naively.

\mypar{Sparse scene graph.}
Instead, we wish to only feed a small but sufficient subset of all possible pairs to \master{}, which structure forms a scene graph $\G$.
Formally, $\G=(\V,\E)$ is a graph where each vertex $I\in\V$ is an image, and each edge $e=(n,m)\in\E$ is an undirected connection between two likely-overlapping images $I^{n}$ and $I^{m}$.
Importantly, $\G$ must have a single connected component, \ie all images must (perhaps indirectly) be linked together.

\mypar{Image retrieval.}
To select the right subset of pairs, we rely on a scalable pairwise image matcher $h(I^{n},I^{m}) \mapsto s$, able to predict the approximate co-visibility score $s\in[0,1]$ between two images $I^{n}$ and $I^{m}$. 
While any off-the-shelf image retriever can in theory do, we propose to leverage \master{}'s encoder $\enc(\cdot)$.
Indeed, our findings are that the encoder, due to its role of laying foundations for the decoder, is implicitly trained for image matching (see~\cref{sub:xpretrieval}).
To that aim, we adopt the ASMK (Aggregated Selective Match Kernels) image retrieval method~\cite{asmk} considering the token features output by the encoder as local features. 
ASMK has shown excellent performance for retrieval, especially without requiring any spatial verification.
In a nutshell, we consider the output $\EF{}$ of the encoder as a bag of local features, apply feature whitening, quantize them according to a codebook previously obtained by k-means clustering, then aggregate and binarize the residuals for each codebook element, thus yielding high-dimensional sparse binary representations.
The ASMK similarity between two image representations can be efficiently computed by summing a small kernel function on binary representations over the common codebook elements.
Note that this method is training-free, only requiring to compute the whitening matrix and the codebook once from a representative set of features. 
We have also try learning a small projector on top of the encoder features following the HOW approach~\cite{how}, but this leads to similar performances. We refer to the supplementary for more details.
The output from the retrieval step is a similarity matrix $S \in [0,1]^{N \times N}$.

\mypar{Graph construction.}
To get a small number of pairs while still ensuring a single connected component, we build the graph $\G$ as follows.
We first select a fixed number $N_a$ of \emph{key images} (or keyframes) using farthest point sampling (FPS)~\cite{fps} based on $S$.
These keyframes constitute the core set of nodes and are densely connected together. 
All remaining images are then connected to their closest keyframe as well as their $k$ nearest neighbors according to $S$. %
Such a graph comprises $O(N_a^2 + (k+1)N) = O(N) \ll O(N^2)$ edges, which is linear in the number of images $N$. 
We typically use $N_a=20$ and $k=10$.
Note that, while the retrieval step has quadratic complexity in theory, it is extremely fast and scalable in practice, so we ignore it in and report quasi-linear complexity overall.

\subsection{Local reconstruction}
\label{sub:localrecon}

As indicated in \cref{sec:prelim}, we run the inference of \master{} for every pair $e=(n,m)\in\E$, yielding raw pointmaps and sparse pixel matches $\M^{n,m}$.
Since MASt3R is order-dependent in terms of its input, we define $\M^{n,m}$ as the union of correspondences obtained by running both $f(\I{n},\I{m})$ and $f(\I{m},\I{n})$. 
Doing so, we also obtain pointmaps $\X{n}{n}, \X{n}{m}, \X{m}{n}$ and $\X{m}{m}$, where $\X{n}{m} \in \R^{H \times W \times 3}$ denotes a 2D-to-3D mapping from pixels of image $\I{n}$ to 3D points in the coordinate system of image $\I{m}$.
Since the encoder features $\{\EF{n}\}_{n=1..N}$ have already been extracted and cached during scene graph construction (\cref{sub:scenegraph}), we only need to run the ViT decoder $\dec()$, which substantially saves time and compute.

\mypar{Canonical pointmaps.}
We wish to estimate an initial depthmap $\depth{n}$ and camera intrinsics $\K{n}$ for each image $\I{n}$.
These can be easily recovered from a raw pointmap $\X{n}{n}$ as demonstrated in~\cite{dust3r}, but note that each pair $(n,\cdot)$ or $(\cdot,n) \in \E$ would yield its own estimate of $\X{n}{n}$.
To average out regression imprecision, we hence aggregate these copycat pointmaps into a canonical pointmap $\cX{n}$. 
Let $\E^{n}=\{e|e \in \E \wedge n \in e\}$ be the set of all edges connected to image $I^{n}$. 
For each edge $e\in\E^{n}$, we have a different estimate of $X^{n,n}$ and its respective confidence maps $C^{n,n}$, which we will denote as $X^{n,e}$ and $C^{n,e}$ in the following. 
We compute the canonical pointmap as a simple per-pixel weighted average of all estimates:
\begin{equation}
\cX{n}_{i,j}=\frac{\sum_{e\in\E^{n}}C_{i,j}^{n,e}X_{i,j}^{n,e}}{\sum_{e\in\E^{n}}C_{i,j}^{n,e}}.
\end{equation}
From it, we then recover the canonical depthmap $\cdepth{n}=\cX{n}_{:,:,3}$ and the focal length using Weiszfeld algorithm~\cite{dust3r}:
{\small
\begin{equation}
f^*=\argmin_{f} \sum_{i,j} \left\Vert \left(i{-}\frac{W}{2},j{-}\frac{H}{2}\right) - f \left(\frac{\tilde{X}^n_{i,j,1}}{\tilde{X}^n_{i,j,3}}, \frac{\tilde{X}^n_{i,j,2}}{\tilde{X}^n_{i,j,3}}\right) \right\Vert,
\end{equation}
}\noindent
which, assuming centered principal point and square pixels, yields the canonical intrinsics $\cK{n}$.
In this work, we assume a pinhole camera model without lens distortion, but our approach could be extended to different camera types.

\mypar{Constrained pointmaps.}
Camera intrinsics $\K{}$, extrinsics $\cam{}$ and depthmaps $\depth{}$ will serve as basic ingredients (or rather, optimization variables) for the global reconstruction phase.
Let $\pi_{n}: \R^3 \mapsto \R^2$ denote the reprojection function onto the camera screen of $\I{n}$, \ie $\pi_{n}(x)=\K{n} \cam{n} \sigma_n x$ for a 3D point $x\in \R^3$ ($\sigma_n>0$ is a per-camera scale factor, \ie we use scaled rigid transformations). 
To ensure that pointmaps perfectly satisfy the pinhole projective model (they are normally over-parameterized), we define a \emph{constrained pointmap} $\chi^n \in \R^{H \times W \times 3}$  explicitly as a function of $\K{n}, \cam{n}, \sigma_n$ and $\depth{n}$.
Formally, the 3D point $\chi^{n}_{i,j}$ seen at pixel $(i,j)$ of image $\I{n}$ is defined using inverse reprojection as $\chi_{i,j}^{n} = \pi_{n}^{-1}(\sigma_n,\K{n},\cam{n},\depth{n}_{i,j}) =  \nicefrac{1}{\sigma_n} \cam{n}^{-1} \K{n}^{-1} \depth{n}_{i,j} \left[i,j,1\right]^{\top}$.

\begin{figure}
    \centering
    \includegraphics[width=0.9\linewidth, trim=0 322 525 0, clip]{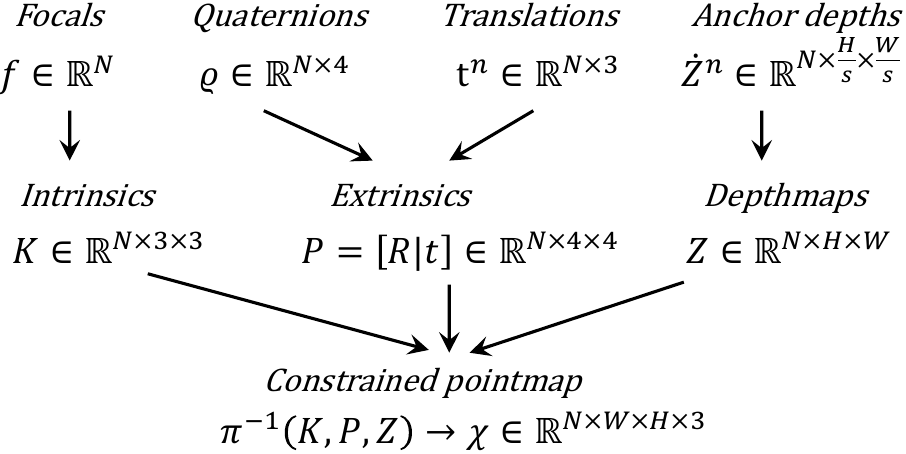}
    \vspace{-0.2cm}
    \caption{\textbf{Factor graph for \ours{}}.
        Free variables on the top row serve to construct the constrained pointmap $\chi$, which follows the pinhole camera model by design and onto which the loss functions from \cref{eq:err3d,eq:err2d} are defined. 
        }
    \vspace{-0.2cm}
    \label{fig:variables}
\end{figure}

\subsection{Coarse alignment}
\label{sub:coarse}

Recently, \duster{}~\cite{dust3r} introduced a global alignment procedure aiming to rigidly move dense pointmaps in a world coordinate system based on pairwise relationships between them.
In this work, we simplify and improve this procedure by taking advantage of pixel correspondences, thereby reducing the overall number of parameters and its memory and computational footprint.

Specifically, we look for the scaled rigid transformations $\sigma^*, \cam{}^*$ of every canonical pointmaps $\chi = \pi^{-1}(\sigma,\cK{},\cam{},\cdepth{})$ (\ie fixing intrinsics $\K{}=\cK{}$ and depth $\depth{}=\cdepth{}$ to their canonical values) such that any pair of matching 3D points gets as close as possible: %
\begin{equation}
\sigma^{*}, P^{*} = \argmin_{\sigma,P} 
        \sum_{\substack{ c\in\mathcal{M}^{n,m} \\ (n,m)\in\mathcal{E} }}   
            q_{c} \left\Vert \chi^{n}_c - \chi^m_c \right\Vert ^ {\lambda_1},
\label{eq:err3d}
\end{equation}
where $c$ denotes the matching pixels in each respective image by a slight abuse of notation.
In contrast to the global alignment procedure in \duster{}, this minimization only applies to sparse pixel correspondences  $y_{c}^{n}\leftrightarrow y_{c}^{m}$ weighted by their respective confidence $q_{c}$ (also output by \master{}).
To avoid degenerate solutions, we enforce $\min_n \sigma_{n}=1$ by reparameterizing $\sigma_n = \sigma'_n / (min_n \sigma'_n)$. %
We minimize this objective using Adam~\cite{adam} for a fixed number $\nu_1$ of iterations.

\subsection{Refinement}
\label{sub:refine}

Coarse alignment converges well and fast in practice, but restricts itself to rigid motion of canonical pointmaps.
Unfortunately, pointmaps are bound to be noisy due to depth ambiguities during local reconstruction.
To further refine cameras and scene geometry, we thus perform a second round of global optimization akin to bundle adjustment~\cite{goos_bundle_2000} with gradient descent for $\nu_2$ iterations and starting from the coarse solution $\sigma^*, \cam{}^*$ obtained from \cref{eq:err3d}.
In other words, we minimize the 2D reprojection error of 3D points in all cameras:
\begin{align}
  & Z^*,K^*,P^*,\sigma^* = \argmin_{Z,K,P,\sigma} \mathcal{L}_2, \text{ with} \label{eq:err2d} \\
  \mathcal{L}_2 = & \sum_{\substack{ c\in\mathcal{M}^{n,m} \\ (n,m)\in\mathcal{E}}}
      q_{c} \left[ \rho\left(y_{c}^{n} - \pi_{n}\left(\chi_{c}^{m}\right) \right)
                 + \rho\left(y_{c}^{m} - \pi_{m}\left(\chi_{c}^{n}\right) \right)\right], \nonumber
\end{align}
with $\rho:\R^2 \mapsto \R^+$ a robust error function able to deal with potential outliers among all extracted correspondences. 
We typically set $\rho(x){=}\left\Vert x\right\Vert ^ {\lambda_2}$ with $0 < \lambda_2 \le 1$ (\eg $\lambda_2{=}0.5$).

\mypar{Forming pseudo-tracks.}
Optimizing \cref{eq:err2d} has little effect, because sparse pixel correspondences $\M^{m,n}$ are rarely \emph{exactly} overlapping across several pairs.
As an illustration, two correspondences $y^m_{\cdot,\cdot} \leftrightarrow y^n_{i,j}$ and $y^n_{i+1,j} \leftrightarrow y^l_{\cdot,\cdot}$ from image pairs $(m,n)$ and $(n,l)$ would independently optimize the two 3D points $\chi^n_{i,j}$ and $\chi^n_{i+1,j}$, possibly moving them very far apart despite this being very unlikely as $(i,j)\simeq (i+1,j)$.
Traditional SfM methods resort to forming point tracks, which is relatively straightforward with keypoint-based matching~\cite{sift,colmapsfm,pixsfm,superglue,superpoint}. 
We propose instead to form pseudo-tracks by creating \emph{anchor points} and rigidly tying together every pixel with their closest anchor point. 
This way, correspondences that do not overlap exactly are still both tied to the same anchor point with a high probability.
Formally, we define anchor points with a regular pixel grid $\dot{y}\in\R^{W/s\times H/s\times 2}$ spaced by $\delta$ pixels:
\begin{equation}
\dot{y}_{u,v} = \left( u\delta+\frac{\delta}{2}, v\delta+\frac{\delta}{2} \right).
\end{equation}
We then tie each pixel $(i,j)$ in $I^{n}$ with its closest anchor $\dot{y}_{u,v}$ at coordinate $(u,v)=(\left\lfloor \nicefrac{i}{\delta} \right\rfloor, \left\lfloor \nicefrac{j}{\delta} \right\rfloor )$.
Concretely, we simply index the depth value at pixel $(i,j)$ to the depth value $\dot{Z}_{u,v}$ of its anchor point, \ie we define $\depth{}_{i,j} = o_{i,j} \dot{Z}_{u,v}$ where $o_{i,j} = \cdepth{}_{i,j} / \cdepth{}_{u,v}$ is a constant relative depth offset calculated at initialization from the canonical depthmap $\cdepth{}$.
Here, we make the assumption that canonical depthmaps are locally accurate. 
All in all, optimizing a depthmap $\depth{n}\in\mathbb{R}^{W\times H}$ thus only comes down to optimizing a reduced set of anchor depth values $\dot{Z}^n \in \mathbb{R}^{W/\delta\times H/\delta}$ (\eg reduced by a factor of 64 if $\delta=8$).

\begin{table*}
    \centering
    \resizebox{0.57\linewidth}{!}{

\setlength{\tabcolsep}{2pt}
\newdimen\wate \wate=30pt
\newdimen\wperc \wperc=24pt
\begin{tabular}{lrrrrrrrrrr}
\toprule
\multirow{2}{*}{Method} & \multicolumn{2}{c}{25 views} & \multicolumn{2}{c}{50 views} & \multicolumn{2}{c}{100 views} & \multicolumn{2}{c}{200 views} & \multicolumn{2}{c}{full} \\
\cmidrule(lr){2-3} \cmidrule(lr){4-5} \cmidrule(lr){6-7} \cmidrule(lr){8-9} \cmidrule(lr){10-11}
 & \multicolumn{1}{c}{ATE$\downarrow$} & \multicolumn{1}{c}{Reg.$\uparrow$} &  \multicolumn{1}{c}{ATE$\downarrow$} & \multicolumn{1}{c}{Reg.$\uparrow$} & \multicolumn{1}{c}{ATE$\downarrow$} & \multicolumn{1}{c}{Reg.$\uparrow$} & \multicolumn{1}{c}{ATE$\downarrow$} & \multicolumn{1}{c}{Reg.$\uparrow$} & \multicolumn{1}{c}{ATE$\downarrow$} & \multicolumn{1}{c}{Reg.$\uparrow$} \\
\midrule
COLMAP~\cite{colmapsfm} & \cellCD[\wate]{c0d6a8}{0.03840} & \cellCD[\wperc]{fee7cb}{44.4} & \cellCD[\wate]{c1d7a8}{0.02920} & \cellCD[\wperc]{fdf0cf}{60.5} & \cellCD[\wate]{fdf2d0}{0.02640} & \cellCD[\wperc]{d2dfb4}{85.7} & \cellCD[\wate]{fee9cc}{0.01880} & \cellCD[\wperc]{b9d3a3}{97.0} & - & - \\
ACE-Zero~\cite{acezero} & \cellCD[\wate]{eac2c2}{0.11160} & \cellCD[\wperc]{b3d09f}{100.0} & \cellCD[\wate]{ecc5c3}{0.07130} & \cellCD[\wperc]{b3d09f}{100.0} & \cellCD[\wate]{f5d2c6}{0.03980} & \cellCD[\wperc]{b3d09f}{100.0} & \cellCD[\wate]{fee9cc}{0.01870} & \cellCD[\wperc]{b3d09f}{100.0} & \cellCD[\wate]{eac2c2}{0.01520} & \cellCD[\wperc]{b3d09f}{100.0} \\
FlowMap~\cite{flowmap} & \cellCD[\wate]{eec7c3}{0.10700} & \cellCD[\wperc]{b3d09f}{100.0} & \cellCD[\wate]{eac2c2}{0.07310} & \cellCD[\wperc]{b3d09f}{100.0} & \cellCD[\wate]{eac2c2}{0.04460} & \cellCD[\wperc]{b3d09f}{100.0} & \cellCD[\wate]{eac2c2}{0.02420} & \cellCD[\wperc]{b3d09f}{100.0} & \cellCD[\wate]{cccccc}{N/A} & \cellCD[\wperc]{fdf3d0}{66.7} \\
VGGSfM~\cite{vggsfm} & \cellCD[\wate]{f9f1cd}{0.05800} & \cellCD[\wperc]{bbd4a4}{96.2} & \cellCD[\wate]{dbe3ba}{0.03460} & \cellCD[\wperc]{b6d1a1}{98.7} & \cellCD[\wate]{feedce}{0.02900} & \cellCD[\wperc]{b6d1a1}{98.5} & \cellCD[\wate]{cccccc}{N/A} & \cellCD[\wperc]{fee9cc}{47.6} & \cellCD[\wate]{cccccc}{N/A} & \cellCD[\wperc]{eac2c2}{0.0} \\
DF-SfM~\cite{detfreeSfm} & \cellCD[\wate]{ffe4ca}{0.08110} & \cellCD[\wperc]{b4d0a0}{99.4} & \cellCD[\wate]{faf2ce}{0.04120} & \cellCD[\wperc]{b3d09f}{100.0} & \cellCD[\wate]{fdf1cf}{0.02710} & \cellCD[\wperc]{b3d09f}{99.9} & \cellCD[\wate]{cccccc}{N/A} & \cellCD[\wperc]{ffe1c9}{33.3} & \cellCD[\wate]{cccccc}{N/A} & \cellCD[\wperc]{e7e9c2}{76.2} \\
\midrule
\textbf{MASt3R-SfM} & \cellCD[\wate]{b3d09f}{0.03360} & \cellCD[\wperc]{b3d09f}{100.0} & \cellCD[\wate]{b3d09f}{0.02610} & \cellCD[\wperc]{b3d09f}{100.0} & \cellCD[\wate]{b3d09f}{0.01680} & \cellCD[\wperc]{b3d09f}{100.0} & \cellCD[\wate]{b3d09f}{0.01300} & \cellCD[\wperc]{b3d09f}{100.0} & \cellCD[\wate]{b3d09f}{0.01060} & \cellCD[\wperc]{b3d09f}{100.0} \\
\bottomrule
\end{tabular}
}
    \hfill
    \resizebox{0.40\linewidth}{!}{\setlength{\tabcolsep}{2pt}
\newdimen\wate \wate=30pt
\begin{tabular}{lrrrr}
\toprule
    Method &        \multicolumn{1}{c}{MIP-360} &           \multicolumn{1}{c}{LLFF} &           \multicolumn{1}{c}{T\&T} &         \multicolumn{1}{c}{CO3Dv2} \\
\midrule
NoPE-NeRF~\cite{nopenerf} & \cellCD[\wate]{eac2c2}{0.04429} & \cellCD[\wate]{eac2c2}{0.03920} & \cellCD[\wate]{eac2c2}{0.03709} & \cellCD[\wate]{eac2c2}{0.03648} \\
DROID-SLAM~\cite{droid} & \cellCD[\wate]{b3d09f}{0.00017} & \cellCD[\wate]{b3d09f}{0.00074} & \cellCD[\wate]{b3d09f}{0.00122} & \cellCD[\wate]{fdf0cf}{0.01728} \\
FlowMap~\cite{flowmap} & \cellCD[\wate]{b5d1a0}{0.00055} & \cellCD[\wate]{bbd4a4}{0.00209} & \cellCD[\wate]{b3d09f}{0.00124} & \cellCD[\wate]{fdf2d0}{0.01589} \\
ACE-Zero~\cite{acezero} & \cellCD[\wate]{bbd4a4}{0.00173} & \cellCD[\wate]{c5d9ab}{0.00396} & \cellCD[\wate]{e8e9c2}{0.00973} & \cellCD[\wate]{b3d09f}{0.00520} \\
\midrule
\textbf{MASt3R-SfM} & \cellCD[\wate]{b6d1a1}{0.00079} & \cellCD[\wate]{b4d0a0}{0.00098} & \cellCD[\wate]{b8d2a2}{0.00215} & \cellCD[\wate]{b4d0a0}{0.00538} \\
\bottomrule
\end{tabular}

}
    \vspace{-0.25cm}
    \caption{\textbf{Results on Tanks\&Temples} in terms of ATE and overall registration rate (Reg.).
        For easier readability, we color-code ATE results as a linear gradient between 
        \BGcolor{e0a4a4}{ w}\BGcolor{e7b2ac}{o}\BGcolor{eec0b5}{r}\BGcolor{f6cebe}{s}\BGcolor{fcdcc6}{t}\BGcolor{ffe4ca}{ }\BGcolor{fee8cc}{a}\BGcolor{feeccd}{n}\BGcolor{fdf0cf}{d}\BGcolor{f6f0cb}{ }\BGcolor{e3e6bd}{b}\BGcolor{cfdcae}{e}\BGcolor{bbd2a0}{s}\BGcolor{a8c992}{t }
        ATE for a given dataset or split; and Reg results with linear gradient between 
        \BGcolor{e0a4a4}{ 0}\BGcolor{e9b6af}{\%}\BGcolor{f3c9ba}{ }\BGcolor{fcdbc5}{a}\BGcolor{ffe5ca}{n}\BGcolor{feeacc}{d}\BGcolor{fdefcf}{ }\BGcolor{f4efc9}{1}\BGcolor{dbe2b7}{0}\BGcolor{c1d5a4}{0}\BGcolor{a8c992}{\%. }
        \textbf{Left}: impact of the number of input views, regularly sampled from the full set.
        `N/A' indicates that at least one scene did not converge.
            \textbf{Right}: ATE$\downarrow$ on different datasets with the arbitrary splits defined in FlowMap~\cite{flowmap}. %
            }
    \vspace{-0.25cm}
    \label{tab:tandt}
\end{table*}

\section{Experimental Results}
\label{sec:expes}

After presenting the datasets and metrics, 
we extensively %
compare our approach with %
state-of-the-art SfM methods in diverse conditions. %
We finally present several ablations. %

\subsection{Experimental setup}

We use the publicly available \master{} checkpoint for our experiments, which we do \emph{not} finetune unless otherwise mentioned.
When building the sparse scene graph in \cref{sub:scenegraph}, we use $N_a=20$ anchor images and $k=10$ non-anchor nearest neighbors.
We use the same grid spacing of $\delta=8$ pixels for extracting sparse correspondences with FastNN (\cref{sub:localrecon}) and defining anchor points (\cref{sub:refine}).
For the two gradient descents, we use the Adam optimizer~\cite{adam} with a learning rate of $0.07$ (resp. $0.014$) for $\nu_1=300$ iterations and $\lambda_1=1.5$ (resp. $\nu_2=300$ and $\lambda_2=0.5$) for the coarse (resp. refinement) optimization, each time with a cosine learning rate schedule and without weight decay.
Unless otherwise mentioned we assume shared intrinsics and optimize a shared per-scene focal parameter for all cameras.

\begin{table}[t]
    \centering
    \resizebox{\linewidth}{!}{\begin{tabular}{llccccc}
\toprule
& \multirow{2}{*}{Method}  & \multicolumn{3}{c}{Co3Dv2$\uparrow$} &  & RealEstate10K$\uparrow$ \\ \cline{3-5} \cline{7-7} %
                       & \hspace{0.1pt} & RRA@15  & RTA@15 & mAA(30) &  & mAA(30)       \\ 
\midrule
\multirow{8}{*}{(a)} & Colmap+SG~\cite{superpoint, superglue} & 36.1    & 27.3   & 25.3    &  & 45.2          \\
&PixSfM~\cite{pixsfm}                  & 33.7    & 32.9   & 30.1    &  & 49.4          \\
&RelPose~\cite{relpose}              & 57.1    & -      & -       &  & -             \\
&PosReg~\cite{posediffusion}           & 53.2    & 49.1   & 45.0    &  & -             \\
&PoseDiff ~\cite{posediffusion}   & 80.5    & 79.8   & 66.5    &  & 48.0          \\
&RelPose++~\cite{relposepp}           & (85.5)    & -      & -       &  & -             \\
&RayDiff~\cite{raydiffusion}   & (93.3)    & -   & -    &  & -          \\ %
&\duster-GA~\cite{dust3r}             & \textbf{96.2}    & 86.8   & 76.7    &  & 67.7          \\ 
&\textbf{\ours}             & 96.0 & \textbf{93.1}   &  \textbf{88.0}   &  & \textbf{86.8} \\ \midrule %

\multirow{2}{*}{(b)} & \duster~\cite{dust3r}               & 94.3    & 88.4  & 77.2    &  & 61.2          \\
&\master{}~\cite{mast3r}            & 94.6   & 91.9   &  81.8 &  & 76.4 \\ 
\bottomrule\\
\end{tabular}
}
    \vspace{-0.55cm}
    \caption{\textbf{Multi-view pose regression on CO3Dv2~\cite{co3d} and 
            RealEstate10K~\cite{realestate10K} with 10 random frames}. 
            Parenthesis () denote methods that do not report results on the 10 views set, we report their best for comparison (8 views). 
            We distinguish between (a) multi-view and (b) pairwise methods. 
       }
    \vspace{-0.4cm}
    \label{tab:relpose_mvs}
\end{table}

\mypar{Datasets.}

To showcase the robustness of our approach, %
we experiment in different conditions representative of diverse experimental setups (video or unordered image collections, simple or complex scenes, outdoor, indoor or object-centric, \etc).
Namely, we employ 
Tanks\&Temples~\cite{tanksandtemples} (T\&T), a 3D reconstruction dataset comprising 21 scenes ranging from 151 to 1106 images; 
ETH3D~\cite{eth3d}, a multi-view stereo dataset with 13 scenes for which ground-truth is available; 
CO3Dv2~\cite{co3d}, an object-centric dataset for multi-view pose estimation; 
and RealEstate10k~\cite{realestate10K}, MIP-360~\cite{mipnerf360} and LLFF~\cite{llff}, three datasets for novel view synthesis.

\mypar{Evaluation metrics.}

For simplicity, we evaluate all methods \wrt ground-truth cameras poses. 
For Tanks\&Temples where it is not provided, we make a pseudo ground-truth with COLMAP~\cite{colmapsfm} using all frames. %
Even though this is not perfect, COLMAP is known to be reliable in conditions where there is a large number of frames with high overlap.
We evaluate the average translation error (ATE) as in FlowMap~\cite{flowmap}, \ie we align estimated camera positions to ground-truth ones with Procrustes~\cite{procrustes} and report an average normalized error. 
We ignore unregistered cameras when doing Procrustes, which favors methods that can reject hard images (such as COLMAP~\cite{colmapsfm} or VGGSfM~\cite{vggsfm}). Note that our method always outputs a pose estimate for all cameras by design, thus negatively impacting our results with this metric.
We also report the relative rotation and translation accuracies (resp. RTA@$\tau$ and RRA@$\tau$, where $\tau$ indicates the threshold in degrees), computed at the pairwise level and averaged over all image pairs~\cite{posediffusion}.
Similarly, the mean Average Accuracy (mAA)@$\tau$ is defined as the area under the curve of the angular differences at min(RRA@$\tau$, RTA@$\tau$).
Finally, we report the successful registration rate as a percentage, denoted as Reg.
When reported at the dataset level, metrics are averaged over all scenes.

\subsection{Comparison with the state of the art}

We first evaluate the impact of the amount of overlap between images on the quality of the SfM output for different state-of-the-art methods.
To that aim, we choose Tanks\&Temple, a standard reconstruction dataset captured with high overlap (originally video frames).
We form new splits by regularly subsampling the original images for 25, 50, 100 and 200 frames.  
Following~\cite{flowmap}, we report results in terms of Average Translation Error (ATE) against the COLMAP pseudo ground-truth in \cref{tab:tandt} (left), computed from the full set of frames and likewise further subsampled.
\ours{} provides nearly constant performance for all ranges, significantly outperforming COLMAP, Ace-Zero, FlowMap and VGGSfM in all settings.
Unsurprisingly, the performance of these methods strongly degrades in small-scale settings (or does not even converge on some scenes for COLMAP). 
On the other hand, we note that FlowMap and VGGSfM crash when dealing with large collections due to insufficient memory despite using 80GB GPUs.

\mypar{FlowMap splits.}
We also report results on the custom splits from the FlowMap paper~\cite{flowmap}, which concerns 3 additional datasets beyond T\&T (LLFF, Mip-360 and CO3Dv2). 
We point out that, not only these splits select a \emph{subset} of scenes for each dataset (in details: 3 scenes from Mip-360, 7 from LLFF, 14 from T\&T and 2 from CO3Dv2), they also select an \emph{arbitrary subset} of consecutive frames in the corresponding scenes. %
Results in \cref{tab:tandt} (right) show that our method is achieving better results than NopeNeRF and ACE-Zero, on par with FlowMap overall and slightly worse than DROID-SLAM~\cite{droid}, a method that can only work in video settings.
Since we largely outperform FlowMap when using regularly sampled splits, we hypothesize that FlowMap is very sensitive to the input setting. %

\mypar{Multi-view pose estimation.}
In \cref{fig:teaser} (top), we also compare to various baselines on CO3Dv2 and RealEstate10K, varying the number of input images by random sampling.
We follow the PoseDiffusion~\cite{posediffusion} splits and protocol for comparison purposes.
We provide detailed comparisons in \cref{tab:relpose_mvs} with state-of-the-art multi-view pose estimation methods, whose goal is only to recover cameras poses but not the scene geometry.
Again, our approach compares favorably to existing methods, particularly when the number of input images is low.
Overall, this highlights that \ours{} is extremely robust to sparse view setups, with its performance not degrading when decreasing the number of views, even for as little as three views. 

\begin{table*}
    \centering
    \resizebox{0.78\linewidth}{!}{
\newdimen\wperc \wperc=24pt
\begin{tabular}{lr@{~}rr@{~}rr@{~}rr@{~}rr@{~}rr@{~}r}
\toprule
       \multirow{2}{*}{Scenes} & \multicolumn{2}{c}{COLMAP~\cite{colmapsfm}} & \multicolumn{2}{c}{ACE-Zero~\cite{acezero}} & \multicolumn{2}{c}{FlowMap~\cite{flowmap}} & \multicolumn{2}{c}{VGGSfM~\cite{vggsfm}} & \multicolumn{2}{c}{DF-SfM~\cite{detfreeSfm}} & \multicolumn{2}{c}{\textbf{MASt3R-SfM}} \\
       \cmidrule(lr){2-3} \cmidrule(lr){4-5}  \cmidrule(lr){6-7}  \cmidrule(lr){8-9} \cmidrule(lr){10-11} \cmidrule(lr){12-13}
              &  \multicolumn{1}{c}{\footnotesize RRA@5} &  \multicolumn{1}{c}{\footnotesize RTA@5} & \multicolumn{1}{c}{\footnotesize RRA@5} & \multicolumn{1}{c}{\footnotesize RTA@5} &  \multicolumn{1}{c}{\footnotesize RRA@5} &  \multicolumn{1}{c}{\footnotesize RTA@5} & \multicolumn{1}{c}{\footnotesize RRA@5} & \multicolumn{1}{c}{\footnotesize RTA@5} & \multicolumn{1}{c}{\footnotesize RRA@5} & \multicolumn{1}{c}{\footnotesize RTA@5} & \multicolumn{1}{c}{\footnotesize RRA@5} & \multicolumn{1}{c}{\footnotesize RTA@5}  \\
\midrule
{\small courtyard} & \cellCD[\wperc]{fceace}{56.3} & \cellCD[\wperc]{dfe1bc}{60.0} & \cellCD[\wperc]{eac2c2}{4.0} & \cellCD[\wperc]{eac2c2}{1.9} & \cellCD[\wperc]{ecc6c3}{7.5} & \cellCD[\wperc]{ebc4c2}{3.6} & \cellCD[\wperc]{fde8cd}{50.5} & \cellCD[\wperc]{f9ebcd}{51.2} & \cellCD[\wperc]{cad9ae}{80.7} & \cellCD[\wperc]{b3d09f}{74.8} & \cellCD[\wperc]{b3d09f}{89.8} & \cellCD[\wperc]{d2dcb3}{64.4} \\
{\small delivery area} & \cellCD[\wperc]{f6d4c6}{34.0} & \cellCD[\wperc]{ffe0c9}{28.1} & \cellCD[\wperc]{efcac4}{27.4} & \cellCD[\wperc]{eac2c2}{1.9} & \cellCD[\wperc]{f2cdc5}{29.4} & \cellCD[\wperc]{fbdbc8}{23.8} & \cellCD[\wperc]{eac2c2}{22.0} & \cellCD[\wperc]{f8d6c7}{19.6} & \cellCD[\wperc]{b5d1a0}{82.5} & \cellCD[\wperc]{b3d09f}{82.0} & \cellCD[\wperc]{b3d09f}{83.1} & \cellCD[\wperc]{b3d09f}{81.8} \\
{\small electro} & \cellCD[\wperc]{fde7cc}{53.3} & \cellCD[\wperc]{fde7cc}{48.5} & \cellCD[\wperc]{f3cfc5}{16.9} & \cellCD[\wperc]{eec9c3}{7.9} & \cellCD[\wperc]{eac2c2}{2.5} & \cellCD[\wperc]{eac2c2}{1.2} & \cellCD[\wperc]{e0e1bc}{79.9} & \cellCD[\wperc]{fdeace}{58.6} & \cellCD[\wperc]{dadfb8}{82.8} & \cellCD[\wperc]{d4ddb4}{81.2} & \cellCD[\wperc]{b3d09f}{100.0} & \cellCD[\wperc]{b3d09f}{95.5} \\
{\small facade} & \cellCD[\wperc]{b3d09f}{92.2} & \cellCD[\wperc]{b3d09f}{90.0} & \cellCD[\wperc]{e6e3c0}{74.5} & \cellCD[\wperc]{fcebcf}{64.1} & \cellCD[\wperc]{eac2c2}{15.7} & \cellCD[\wperc]{eac2c2}{16.8} & \cellCD[\wperc]{fde8cd}{57.5} & \cellCD[\wperc]{fee4cb}{48.7} & \cellCD[\wperc]{d3dcb4}{80.9} & \cellCD[\wperc]{c8d8ad}{82.6} & \cellCD[\wperc]{e6e3c0}{74.3} & \cellCD[\wperc]{dfe1bc}{75.3} \\
{\small kicker} & \cellCD[\wperc]{cedbb1}{87.3} & \cellCD[\wperc]{d1dcb3}{86.2} & \cellCD[\wperc]{fad9c7}{26.2} & \cellCD[\wperc]{f4d0c5}{16.8} & \cellCD[\wperc]{eac2c2}{1.5} & \cellCD[\wperc]{eac2c2}{1.5} & \cellCD[\wperc]{b3d09f}{100.0} & \cellCD[\wperc]{b7d2a2}{97.8} & \cellCD[\wperc]{c1d5a8}{93.5} & \cellCD[\wperc]{c7d8ac}{91.0} & \cellCD[\wperc]{b3d09f}{100.0} & \cellCD[\wperc]{b3d09f}{100.0} \\
{\small meadow} & \cellCD[\wperc]{eac2c2}{0.9} & \cellCD[\wperc]{eac2c2}{0.9} & \cellCD[\wperc]{ecc5c3}{3.8} & \cellCD[\wperc]{eac2c2}{0.9} & \cellCD[\wperc]{ecc5c3}{3.8} & \cellCD[\wperc]{ebc4c2}{2.9} & \cellCD[\wperc]{b3d09f}{100.0} & \cellCD[\wperc]{b3d09f}{96.2} & \cellCD[\wperc]{fde8cd}{56.2} & \cellCD[\wperc]{fdeace}{58.1} & \cellCD[\wperc]{fde9cd}{58.1} & \cellCD[\wperc]{fdeace}{58.1} \\
{\small office} & \cellCD[\wperc]{ffe2ca}{36.9} & \cellCD[\wperc]{ffe1c9}{32.3} & \cellCD[\wperc]{eac2c2}{0.9} & \cellCD[\wperc]{eac2c2}{0.0} & \cellCD[\wperc]{eac2c2}{0.9} & \cellCD[\wperc]{ebc3c2}{1.5} & \cellCD[\wperc]{fcebcf}{64.9} & \cellCD[\wperc]{fee4cb}{42.1} & \cellCD[\wperc]{f3e8c9}{71.1} & \cellCD[\wperc]{fde8cd}{54.5} & \cellCD[\wperc]{b3d09f}{100.0} & \cellCD[\wperc]{b3d09f}{98.5} \\
{\small pipes} & \cellCD[\wperc]{fadac7}{30.8} & \cellCD[\wperc]{fcdcc8}{28.6} & \cellCD[\wperc]{ecc5c3}{9.9} & \cellCD[\wperc]{eac2c2}{1.1} & \cellCD[\wperc]{eac2c2}{6.6} & \cellCD[\wperc]{f1ccc4}{12.1} & \cellCD[\wperc]{b3d09f}{100.0} & \cellCD[\wperc]{b7d2a2}{97.8} & \cellCD[\wperc]{f3e9c9}{72.5} & \cellCD[\wperc]{fceace}{61.5} & \cellCD[\wperc]{b3d09f}{100.0} & \cellCD[\wperc]{b3d09f}{100.0} \\
{\small playground} & \cellCD[\wperc]{f3d0c5}{17.2} & \cellCD[\wperc]{f5d2c6}{18.1} & \cellCD[\wperc]{ebc3c2}{3.8} & \cellCD[\wperc]{eac2c2}{2.6} & \cellCD[\wperc]{eac2c2}{2.6} & \cellCD[\wperc]{eac2c2}{2.8} & \cellCD[\wperc]{ffe2c9}{37.3} & \cellCD[\wperc]{fee4cb}{40.8} & \cellCD[\wperc]{f5e9ca}{70.5} & \cellCD[\wperc]{ebe5c4}{70.1} & \cellCD[\wperc]{b3d09f}{100.0} & \cellCD[\wperc]{b3d09f}{93.6} \\
{\small relief} & \cellCD[\wperc]{f6d4c6}{16.8} & \cellCD[\wperc]{f5d3c6}{16.8} & \cellCD[\wperc]{f6d4c6}{16.8} & \cellCD[\wperc]{f6d3c6}{17.0} & \cellCD[\wperc]{eac2c2}{6.9} & \cellCD[\wperc]{eac2c2}{7.7} & \cellCD[\wperc]{b3d09f}{59.6} & \cellCD[\wperc]{b3d09f}{57.9} & \cellCD[\wperc]{fee6cc}{32.9} & \cellCD[\wperc]{fde7cc}{32.9} & \cellCD[\wperc]{fde7cc}{34.2} & \cellCD[\wperc]{fcebcf}{40.2} \\
{\small relief 2} & \cellCD[\wperc]{eec9c3}{11.8} & \cellCD[\wperc]{f2cdc5}{11.8} & \cellCD[\wperc]{eac2c2}{7.3} & \cellCD[\wperc]{ecc5c3}{5.6} & \cellCD[\wperc]{ebc3c2}{8.4} & \cellCD[\wperc]{eac2c2}{2.8} & \cellCD[\wperc]{b3d09f}{69.9} & \cellCD[\wperc]{c4d7aa}{70.3} & \cellCD[\wperc]{fde8cd}{40.9} & \cellCD[\wperc]{fee6cc}{39.1} & \cellCD[\wperc]{dfe1bc}{57.4} & \cellCD[\wperc]{b3d09f}{76.1} \\
{\small terrace} & \cellCD[\wperc]{b3d09f}{100.0} & \cellCD[\wperc]{b3d09f}{100.0} & \cellCD[\wperc]{eac2c2}{5.5} & \cellCD[\wperc]{eac2c2}{2.0} & \cellCD[\wperc]{fcddc8}{33.2} & \cellCD[\wperc]{f8d7c7}{24.1} & \cellCD[\wperc]{ffe2c9}{38.7} & \cellCD[\wperc]{fcdcc8}{29.6} & \cellCD[\wperc]{b3d09f}{100.0} & \cellCD[\wperc]{b4d0a0}{99.6} & \cellCD[\wperc]{b3d09f}{100.0} & \cellCD[\wperc]{b3d09f}{100.0} \\
{\small terrains} & \cellCD[\wperc]{b3d09f}{100.0} & \cellCD[\wperc]{b3d09f}{99.5} & \cellCD[\wperc]{ecc6c3}{15.8} & \cellCD[\wperc]{eac2c2}{4.5} & \cellCD[\wperc]{eac2c2}{12.3} & \cellCD[\wperc]{f0cbc4}{13.8} & \cellCD[\wperc]{fceccf}{70.4} & \cellCD[\wperc]{fde7cd}{54.9} & \cellCD[\wperc]{b3d09f}{100.0} & \cellCD[\wperc]{c4d7aa}{91.9} & \cellCD[\wperc]{fde7cc}{58.2} & \cellCD[\wperc]{fde7cc}{52.5} \\
\midrule
Average & \cellCD[\wperc]{fde8cd}{49.0} & \cellCD[\wperc]{fde8cd}{47.8} & \cellCD[\wperc]{efcac4}{16.4} & \cellCD[\wperc]{ebc3c2}{9.7} & \cellCD[\wperc]{eac2c2}{10.1} & \cellCD[\wperc]{eac2c2}{8.8} & \cellCD[\wperc]{e3e2bf}{65.4} & \cellCD[\wperc]{f3e9c9}{58.9} & \cellCD[\wperc]{c8d8ad}{74.2} & \cellCD[\wperc]{cedbb1}{70.7} & \cellCD[\wperc]{b3d09f}{81.2} & \cellCD[\wperc]{b3d09f}{79.7} \\
\bottomrule
\end{tabular}

}
    \vspace{-0.25cm}
    \caption{\textbf{Detailed per-scene translation and rotation accuracies~($\uparrow$) on ETH-3D.} 
        For clarity, we color-code results with a linear gradient between the
        \BGcolor{e0a4a4}{ w}\BGcolor{e7b2ac}{o}\BGcolor{eec0b5}{r}\BGcolor{f6cebe}{s}\BGcolor{fcdcc6}{t}\BGcolor{ffe4ca}{ }\BGcolor{fee8cc}{a}\BGcolor{feeccd}{n}\BGcolor{fdf0cf}{d}\BGcolor{f6f0cb}{ }\BGcolor{e3e6bd}{b}\BGcolor{cfdcae}{e}\BGcolor{bbd2a0}{s}\BGcolor{a8c992}{t }
        result for a given scene.
    }
    \vspace{-0.25cm}
    \label{tab:eth3d}
\end{table*}

\mypar{Unordered collections.}
We note that benchmarks in previous experiments were originally acquired using video cameras, and then subsampled into frames. 
This might introduce biases that may not well represent the general case of unconstrained SfM. 
We thus experiment on the ETH3D dataset, a photograph dataset, composed of 13 scenes with up to to 76 images per scene. 
Results reported in \cref{tab:eth3d} shows that \ours{} outperforms all competing approaches by a large margin on average.
This is not surprising, as neither ACE-Zero nor FlowMap can handle non-video setups.
The fact that COLMAP and VGGSfM also perform relatively poorly indicates a high sensitivity to not having highly overlapping images, meaning that in the end these methods cannot really handle truly unconstrained collections, in spite of some opposite claims~\cite{vggsfm}.

\subsection{Ablations}

\begin{table}[t]
    \centering
    \resizebox{\linewidth}{!}{\begin{tabular}{lrrrrrr}
\toprule
Scene Graph & \multicolumn{1}{c}{ATE$\downarrow$} & \multicolumn{1}{c}{RTA@5$\uparrow$} & \multicolumn{1}{c}{RRA@5$\uparrow$} & \multicolumn{1}{c}{\#Pairs} & \multicolumn{1}{c}{GPU MEM} & \multicolumn{1}{c}{Avg. T} \\
\midrule
Complete & 0.01256 & 75.9 & 74.8 & 39,800 & 29.9 GB & 2.2 h \\
Local window & 0.02509 & 33.1 & 28.8 & 2,744 & 7.6 GB & 14.1 min \\
Random & 0.01558 & 55.2 & 48.8 & 2,754 & 6.9 GB & 14.7 min \\
\bf{Retrieval} & 0.01243 & 70.9 & 67.6 & 2,758 & 8.4 GB & 14.3 min \\
\bottomrule
\end{tabular}
}
    \vspace{-0.25cm}
    \caption{\textbf{Ablation of scene graph construction} on Tanks\&Temples (200 view subset). See text for details.}
    \vspace{-0.1cm}
    \label{tab:abl-scenegraph}
\end{table}

\begin{table}[t]
    \centering
    \resizebox{\linewidth}{!}{\begin{tabular}{rlcccr}
\toprule
\multicolumn{2}{c}{Ablation} & ATE$\downarrow$ & RTA@5$\uparrow$ & RRA@5$\uparrow$ & \multicolumn{1}{c}{\#Pairs} \\
\midrule
\multirow[m]{3}{*}{Retrieval} & kNN & 0.01440 & 64.1 & 61.9 & 3,042 \\
 & Keyframes & 0.01722 & 58.1 & 57.1 & 740 \\
 & \bf{Keyframes + kNN} & \textbf{0.01243} & \textbf{70.9} & \textbf{67.6} & 2,758 \\
\midrule
\multirow[m]{3}{*}{Optimization level} & Coarse & 0.01504 & 47.4 & 57.7 & 2,758 \\
 & Fine (w/o depth) & 0.01315 & 67.3 & 66.9 & 2758 \\
 & \bf{Fine} & \textbf{0.01243} & \textbf{70.9} & \textbf{67.6} & 2,758 \\
\midrule
\multirow[m]{2}{*}{Intrinsics} & Separate & 0.01329 & 66.9 & 64.2 & 2,758 \\
 & \bf{Shared} & \textbf{0.01243} & \textbf{70.9} & \textbf{67.6} & 2,758 \\
\bottomrule
\end{tabular}
}
    \vspace{-0.25cm}
    \caption{\textbf{Ablations on Tanks\&Temples} (200 view subset). See text for details.}
    \vspace{-0.1cm}
    \label{tab:abl}
\end{table}

We now study the impact of various design choices. %
All experiments are conducted on the Tanks\&Temples dataset regularly subsampled for 200 views per scene. %

\mypar{Scene graph.}
We evaluate different construction strategies for the scene graph in \cref{tab:abl-scenegraph}: `complete' means that we extract all pairs, `local window` is an heuristic for video-based collections that connects every frame with its neighboring frames, and `random' means that we sample random pairs. 
Except for the `complete' case, we try to match the number of pairs used in the baseline retrieval strategy. 
Slightly better results are achieved with the complete graph, but it is about 10x slower than retrieval-based graph and no scalable in general.
Assuming we use retrieval, we further ablate the scene graph building strategy from the similarity matrix in \cref{tab:abl}.
As a reminder, it consists of building a small but complete graph of keyframes, and then connecting each image with the closest keyframe and with $k$ nearest non-keyframes.
We experiment with using only k-NN with an increased $k=13$ to compensate for the missing edges, denoted as `k-NN', or to only use the keyframe graph (\ie $k=0$), denoted as `Keyframe'.
Overall, we find that combining short-range ($k$-NN) and long-range (keyframes) connections is important for reaching top performance.

\mypar{Retrieval with \master{}.}
\label{sub:xpretrieval}
To better assert the effectiveness of our image retrieval strategy alone, we conduct experiments for the task of retrieval-assisted visual localization.
We follow the protocol from~\cite{mast3r} and retrieve the top-$k$ posed images in the database for each query, extract 2D-3D corresponds and run RANSAC to obtain predicted camera poses. 
We compare ASMK on MASt3R features to the off-the-shelf retrieval method FIRe~\cite{fire}, also based on ASMK, on the Aachen-Day-Night~\cite{aachen} and InLoc~\cite{inloc} datasets.
We report standard visual localization accuracy metrics, \ie the percentages of images successfully localized within error thresholds of (0.25m, 2\textdegree) / (0.5m, 5\textdegree) / (5m, 10\textdegree) and (0.25m, 2\textdegree) / (0.5m, 10\textdegree) / (1m, 10\textdegree) respectively.\footnote{\url{https://www.visuallocalization.net/}} in \cref{tab:retrieval}.
Interestingly, using frozen MASt3R features for retrieval performs on par with FIRE, a state-of-the-art method specifically trained for image retrieval and operating on multi-scale features (bottom row).
Our method also reaches competitive performance compared to dedicated visual localization pipelines (top rows), even setting a new state of the art for InLoc.
We refer to the supplementary material for further comparisons. %

\begin{table}
    \centering
    \resizebox{\linewidth}{!}{\begin{tabular}{l@{~~~~}c@{~~~}c@{~~~~~~}c@{~~~}c}
\toprule
\multirow{2}{*}{Method} & \multicolumn{2}{c}{Aachen-Day-Night$\uparrow$} & \multicolumn{2}{c}{InLoc$\uparrow$} \\
\cmidrule(lr){2-3} \cmidrule(lr){4-5} 
& Day & Night & DUC1 & DUC2 \\
\midrule
{\small Kapture~\cite{kapture}+R2D2~\cite{r2d2}} & \textbf{91.3}/\textbf{97.0}/99.5 & \textbf{78.5}/91.6/\textbf{100} & 41.4/60.1/73.7 & 47.3/67.2/73.3 \\
{\small SuperPoint~\cite{superpoint}+LightGlue~\cite{lightglue}} & 90.2/96.0/99.4 & 77.0/91.1/\textbf{100} & 49.0/68.2/79.3 & 55.0/74.8/79.4 \\
{\small LoFTR~\cite{loftr}} & 88.7/95.6/99.0 & \textbf{78.5}/90.6/99.0 & 47.5/72.2/84.8 & 54.2/74.8/85.5 \\ 
{\small DKM~\cite{dkm}} & - & - & 51.5/75.3/86.9 & 63.4/82.4/87.8 \\
\midrule
{\small MASt3R (FIRe top20)} & 89.8/96.8/\textbf{99.6} & 75.9/\textbf{92.7}/\textbf{100} & \textbf{60.6}/\textbf{83.3}/93.4 & 65.6/86.3/88.5 \\
{\small \textbf{MASt3R (MASt3R-ASMK top20)}} & 88.7/94.9/98.2 & 77.5/90.6/97.9 & 58.1/82.8/\textbf{94.4} & \textbf{69.5}/\textbf{90.8}/\textbf{92.4} \\
\bottomrule
\end{tabular}
}
    \vspace{-0.25cm}
    \caption{\textbf{Comparison of retrieval based on MASt3R features } using ASMK with the state-of-the-art FIRe method when localizing with MASt3R (bottom rows), as well as with other state-of-the-art visual localization methods (top rows).}
    \vspace{-0.25cm}
    \label{tab:retrieval}
\end{table}

\begin{figure}
    \centering
    \vspace{-0.15cm}
    \includegraphics[width=\linewidth]{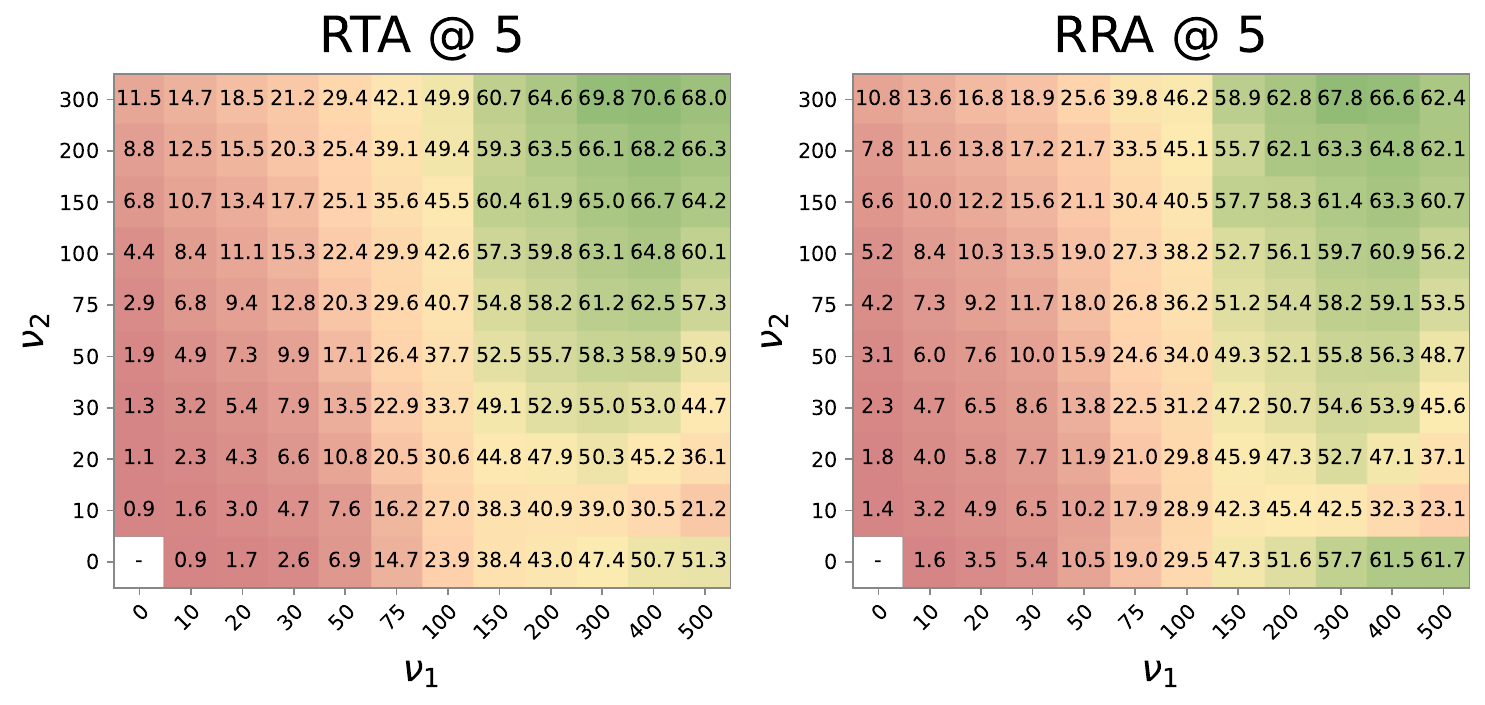}
    \vspace{-0.85cm}
    \caption{\textbf{Pose accuracy ($\uparrow$)} %
    on T\&T-200 \wrt the number of iterations of the coarse and refinement stages (resp. $\nu_1$ and $\nu_2$).}
    \vspace{-0.2cm}
    \label{fig:abl-niter}
\end{figure}

\mypar{Optimization level.}
We also study the impact of the coarse optimization and refinement (\cref{tab:abl}).
As expected, coarse optimization alone, which is somewhat comparable to the global alignment proposed in \duster{} (except we are using sparse matches and less optimization variables), yields significantly less precise pose estimates.
In \cref{fig:abl-niter}, we plot the pose accuracy as a function of the number of iterations during coarse optimization and refinement. 
As expected, refinement, a strongly non-convex bundle-adjustment problem, cannot recover from a random initialization ($\nu_1=0$). Good enough poses are typically obtained after $\nu_1\simeq250$ iterations of coarse optimization, from which point refinement consistently improves. 
We also try to perform the optimization without optimizing depth (\ie using frozen canonical depthmaps, which proves useful for purely rotational cases, denoted as `Fine without depth' in \cref{tab:abl}), in which case we observe a smaller impact on the performance, indicating the high-quality of canonical depthmaps output by \master{} (\cref{sub:localrecon}).

\mypar{Shared intrinsics.}
We finally evaluate the impact of only optimizing one set of intrinsics for all views (`shared'), which is small, indicating that our method is not sensitive to varying intrinsics.

\section{Conclusion}
\label{sec:conclusion}

We have introduced \ours{}, a comparatively simpler fully-integrated solution for unconstrained SfM.
In contrast with current existing SfM pipelines, it can handle very small image collections without apparent issues.
Thanks to the strong priors encoded in the underlying \master{} foundation model upon which our approach is based, it can even deal with cases without motion, and does not rely at all on RANSAC, both features that are normally not possible with standard triangulation-based SfM.

\clearpage
\LARGE
\textbf{Appendix}
\vspace{1cm}
\normalsize

\appendix

\section{Qualitative Results}
\label{app:sup_quali}

\begin{figure*}
    \centering
    \includegraphics[width=.95\linewidth]{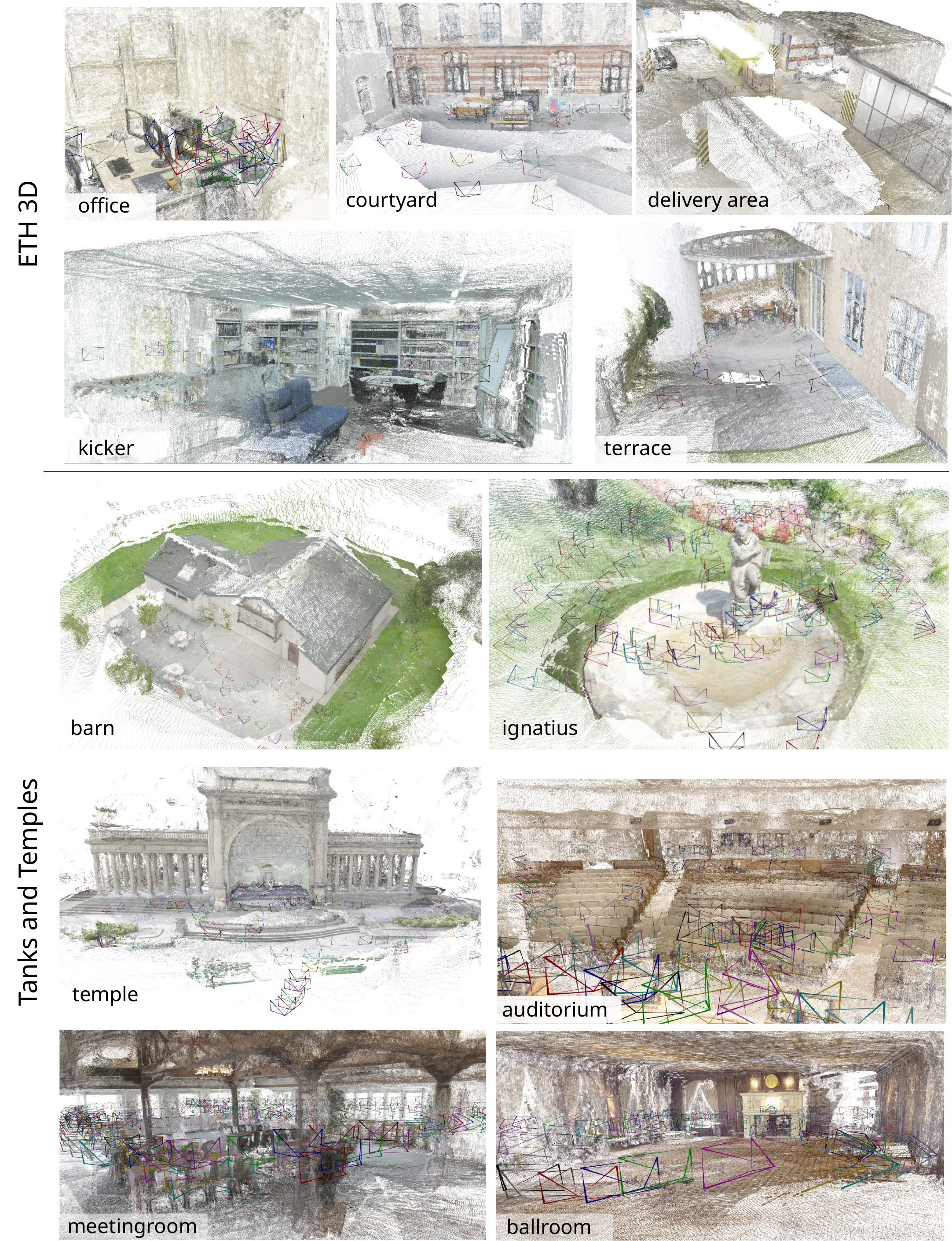}
    \caption{Qualitative reconstruction results for \ours{} on ETH-3D (top) and Tanks\&Temples (bottom). 
    These are the raw outputs of the proposed SfM pipeline, without further refinement.}
    \label{fig:qual}
\end{figure*}

We first present some qualitative reconstruction examples in \cref{fig:qual}.
These are the raw outputs of the proposed SfM pipeline, without further refinement.
We point out that our method produces relatively dense outputs, despite the fact that it only leverages sparse matches.
This is because the inverse reprojection function $\pi^{-1}(\cdot)$ (Section 4.2 of the main paper) can be used to infer a 3D point for \emph{every} pixel, \ie not just those belonging to sparse matches.
Since \master{} is limited to image downscaled to 512 pixels in their largest dimension, we can typically produce about 200K 3D points per image.

\section{Other retrieval variants based on MASt3R features}
\label{app:retrieval}

In the main paper, we propose to use ASMK~\cite{asmk} on the token features output from the MASt3R encoder, after applying whitening.
In this supplementary material, we compare this strategy to using a global descriptor representation per image with a cosine similarity between image representations.
We also compare to a strategy where a small projector is learned on top of the frozen MASt3R encoder feature with ASMK, following an approach similar to HOW~\cite{how} and FIRe~\cite{fire} for training it.
Results are reported in Table~\ref{tab:retrieval2}.

For the global representation, we experimentally find that global average pooling performs slightly better than global max-pooling, and that applying PCA-whitening was beneficial and report this approach. However, the performance of such an approach remains lower than applying ASMK on the token features (top row).

For learning a projector prior to applying ASMK, we follow the strategy of HOW and FIRe, which show that a model can be trained with a standard global representation obtained by a weighted sum of local features. As training dataset, we use the same training data as MASt3R, compute the overlap in terms of 3D points between these image pairs, and consider as positive pairs any pair with more than 10\% overlap, and as negatives pairs coming from two different sequences or datasets. While we observe an improvement in terms of the retrieval mean-average-precision metric on an held-out validation set, this does not yield significant gains when applied to visual localization (bottom row). We thus keep the training-free ASMK approach for \ours.

\begin{table}
    \centering
    \resizebox{\linewidth}{!}{
    \begin{tabular}{l@{~~~~}c@{~~~}c@{~~~~~~}c@{~~~}c}
    \toprule
    \multirow{2}{*}{Retrieval} & \multicolumn{2}{c}{Aachen-Day-Night} & \multicolumn{2}{c}{InLoc} \\
    \cmidrule(lr){2-3} \cmidrule(lr){4-5} 
    & Day & Night & DUC1 & DUC2 \\
    \midrule

    \textbf{MASt3R-ASMK} & \textbf{88.7}/\textbf{94.9}/\textbf{98.2} & \textbf{77.5}/\textbf{90.6}/\textbf{97.9} & 58.1/\textbf{82.8}/\textbf{94.4} & 69.5/90.8/92.4 \\
    MASt3R-global & 86.7/93.7/97.6 & 68.6/84.8/93.2 & \textbf{60.6}/81.8/91.9 & 66.4/87.8/90.8 \\
    MASt3R-proj-ASMK & 88.0/94.8/\textbf{98.2} & 70.2/88.0/94.2 & 60.1/80.8/91.4	& \textbf{74.0}/\textbf{92.4}/\textbf{93.1} \\
    \bottomrule
    \end{tabular}
    }
    \vspace{-0.25cm}
    \caption{\textbf{Comparison of retrieval based on MASt3R features.} We compare the visual localization accuracy using top-20 retrieved images with ASMK (top row), a global feature representation obtained by averaging pooling the local features, whitening using a cosine similarity (middle row), and ASMK when first learning a projector on top of the MASt3R features (bottom row).}
    \label{tab:retrieval2}
\end{table}

\section{Robustness to pure rotations}

We perform additional experiments regarding purely rotational cases, \ie situations where all cameras share the same optical center.
In such cases, the triangulation step from traditional SfM pipeline becomes ill-defined and notoriously fails.
To that aim, we leverage mapping images from the InLoc dataset~\cite{inloc} which are conveniently generated as perspective crops (with a $60^{\circ}$ field-of-view) of 360 panoramic images at three different pitch values, regularly sampled every $30^{\circ}$.
This leads to bundles of 36 RGB images that exactly share a common optical center. 
Using regular sampling, we select 20 sequences from the DUC1 and DUC2 sets and use them to evaluate rotation estimation accuracy.
Results in terms of RRA@5 in \cref{tab:sup-pure-rot} clearly confirm that methods based on the traditional SfM pipeline such as COLMAP~\cite{colmapmvs} or VGGSfM~\cite{vggsfm} do dramatically fail in such a situation.
In contrast, \ours{} performs much better, achieving 100\% accuracy on some scenes, even though it also fail in a few cases.
Disabling the optimization of anchor depth values (\ie fixing depth to the canonical depthmaps) slightly improves the performance.

\begin{figure*}
    \centering
    \includegraphics[width=0.8\linewidth]{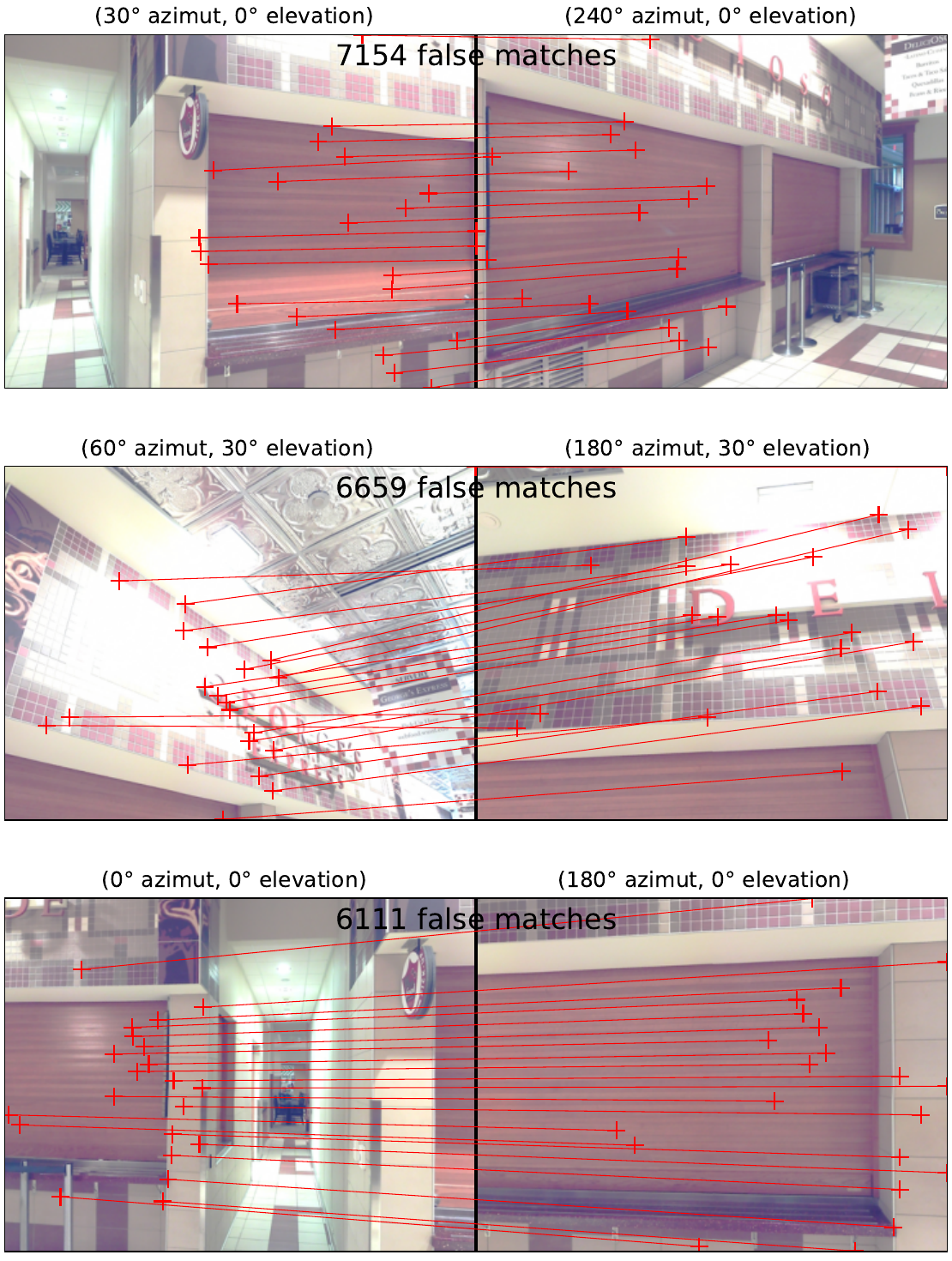}
    \caption{In all failure cases that we have manually reviewed, the root cause of failure was the presence of wrong matches (outliers) between similar-looking parts of the same scene. 
    Here, we show 3 such wrong pairs for the InLoc dataset (purely rotational case, specifically for the scene \texttt{DUC1/007}), each time printing the ground-truth cameras' azimuth and elevation and a small number of randomly-selected matches (showing all of them would impair readibility).
    }
    \label{fig:failure-example}
\end{figure*}

\mypar{Failure cases.}
After analyzing the results, we observe that failures are due to the presence of outlier (false) matches between similar-looking structures.
A few examples of such wrong matching are given in \cref{fig:failure-example}.
These are typically hard outliers that would pass geometric verification.
In fact, the matching problem in such cases becomes ill-defined, since even for a human observer it can be challenging to notice that the two images show different parts of the scene.

\begin{table*}[t]
    \centering
    \resizebox{0.99\linewidth}{!}{\begin{tabular}{lcccccccccccccccccccc|c}
\toprule

Method & \rotatebox{90}{DUC1/000} & \rotatebox{90}{DUC1/007} & \rotatebox{90}{DUC1/014} & \rotatebox{90}{DUC1/021} & \rotatebox{90}{DUC1/070} & \rotatebox{90}{DUC1/077} & \rotatebox{90}{DUC1/084} & \rotatebox{90}{DUC1/091} & \rotatebox{90}{DUC2/033} & \rotatebox{90}{DUC2/040} & \rotatebox{90}{DUC2/047} & \rotatebox{90}{DUC2/054} & \rotatebox{90}{DUC2/061} & \rotatebox{90}{DUC2/093} & \rotatebox{90}{DUC2/100} & \rotatebox{90}{DUC2/107} & \rotatebox{90}{DUC2/115} & \rotatebox{90}{DUC2/122} & \rotatebox{90}{DUC2/129} & \rotatebox{90}{DUC2/132} & Mean \\
\midrule

COLMAP ~\cite{colmapsfm}
&1.0&6.0&4.4&0.5&12.4&0.5&4.4&1.0&1.0&0.5&1.0&2.4&14.4&5.7&7.8&8.4&5.7&0.5&1.3&3.7&4.1\\

FlowMap~\cite{flowmap} & 0.3 &0.2&0.0&0.2&0.0&0.3&0.0&0.0&0.0&0.2&0.0&0.2&0.0&0.0&0.2&0.0&0.2&0.0&0.2&0.0&0.1 \\

VGGSfM~\cite{vggsfm} & 2.5 & 0.0 & 1.0 & 0.5 & 0.0 & 1.0 & 0.0 & 0.2 & 2.1 & 0.0 & 0.0 & 0.0 & 2.9 & 4.1 & 4.9 & 0.3 & 1.0 & 1.1 & 3.3 & 1.6 & 1.3 \\

ACE-Zero~\cite{acezero} & 100.0&100.0&100.0&100.0&100.0&100.0&100.0&100.0&100.0&100.0&100.0&100.0&100.0&100.0&100.0&100.0&100.0&89.0&100.0&100.0&99.5 \\

\midrule

\textbf{MASt3R-SfM}  & 89.0 & 0.8 & 100.0 & 94.4 & 89.0 & 94.4 & 15.1 & 94.6 & 87.5 & 28.7 & 100.0 & 12.9 & 24.8 & 48.3 & 11.0 & 89.0 & 94.4 & 19.0 & 100.0 & 51.0 & 62.2 \\

\textbf{MASt3R-SfM}${}^\dagger$ & 94.4 & 15.2 & 99.5 & 100.0 & 89.0 & 94.4 & 84.0 & 94.4 & 94.4 & 25.1 & 94.4 & 23.0 & 29.7 & 100.0 & 30.5 & 94.4 & 22.2 & 23.5 & 89.0 & 37.1 & 66.7\\

\bottomrule
\end{tabular}

}
    \vspace{-0.25cm}
    \caption{\textbf{Pure Rotation Case.} RRA@5 ($\uparrow$) on 20 randomly chosen scenes from the InLoc dataset. \textbf{MASt3R-SfM}${}^\dagger$ denotes our approach with disabled depth optimization for better optimization stability.}
    \label{tab:sup-pure-rot}
\end{table*}

\section{Additional Results}

\begin{table}[h] 
    \centering
    \resizebox{1\linewidth}{!}{\begin{tabular}{lcccccc}
\toprule
\multirow{2}{*}{Methods} & \multirow{2}{*}{N Frames} & \multicolumn{3}{c}{Co3Dv2~\cite{co3d}} &  & RealEstate10K~\cite{realestate10K}\\ \cline{3-5} \cline{7-7}
                       &  & RRA@15  & RTA@15 & mAA(30) &  & mAA(30)       \\ 
\midrule
COLMAP+SPSG     & 3 &  $\sim$22  &  $\sim$14    &  $\sim$15    &  &    $\sim$23        \\
PixSfM     & 3 &  $\sim$18  &  $\sim$8    &  $\sim$10    &  &    $\sim$17        \\
Relpose     & 3 &  $\sim$56  &  -    &  -    &  &    -        \\
PoseDiffusion     & 3 &  $\sim$75  &  $\sim$75    &  $\sim$61    &  &   -\ ($\sim$77)        \\
VGGSfM     & 3 &  58.7  &    51.2 &  45.4  &  &     -  \\
\duster      & 3 & \textbf{95.3}    & 88.3   & 77.5    &  & 69.5         \\
\textbf{\ours}      & 3 &  94.7   &  \textbf{92.1}  & \textbf{ 85.7}   &  &   \textbf{ 84.3}   \\
\midrule
COLMAP+SPSG     & 5 &  $\sim$21  &  $\sim$17    &  $\sim$17    &  &    $\sim$34        \\
PixSfM     & 5 &  $\sim$21  &  $\sim$16    &  $\sim$15    &  &    $\sim$30        \\
Relpose     & 5 &  $\sim$56  &  -    &  -    &  &    -        \\
PoseDiffusion     & 5 &  $\sim$77  &  $\sim$76    &  $\sim$63    &  &    -\ ($\sim$78)        \\
VGGSfM     & 5 &  80.4  &  75.0   & 69.0  &   & -      \\
\duster      & 5 & \textbf{95.5}    & 86.7   & 76.5    &  & 67.4          \\
\textbf{\ours}      & 5 &   95.0  & \textbf{91.9}   &   \textbf{86.4}  &  &   \textbf{85.3}       \\
\midrule
COLMAP+SPSG     & 10 &  31.6  &  27.3    &  25.3    &  &    45.2        \\
PixSfM     & 10 &  33.7  &  32.9    &  30.1    &  &   49.4        \\
Relpose     & 10 &  57.1  &  -    &  -    &  &    -        \\
PoseDiffusion     & 10 &  80.5  &  79.8    &  66.5    &  &    48.0 ($\sim$80)        \\
VGGSfM     & 10 &  91.5  &  86.8   &  81.9 &  &  -     \\
\duster      & 10 & \bf{96.2}    & 86.8   & 76.7    &  & 67.7          \\ 
\textbf{\ours}      & 10 &  96.0   &  \textbf{93.1}  &  \textbf{88.0}   &  &    \textbf{86.8}      \\ 
\bottomrule
\end{tabular}
}
    \caption{\textbf{Comparison with the state of the art for multi-view pose regression on the CO3Dv2~\cite{co3d} and RealEstate10K~\cite{realestate10K} datasets with 3, 5 and 10 random frames.}
    (Parentheses) indicates results obtained after training on RealEstate10K.
    In contrast, we report results \emph{without} training on RealEstate10K. 
    }
    \label{tab:relpose_mvs2}
\end{table}

\mypar{More comparisons on CO3D and RealEstate10K.}
We provide comparisons with further baselines on the CO3D and RealEstate10K datasets for the cases of 3, 5 and 10 input images in \cref{tab:relpose_mvs2}.
We observe that \ours{} largely outperforms all competing approaches, only neared by DUSt3R which is much less precise overall.

\mypar{Detailed Tanks\&Temple results.}
For completeness, we provide detailed results for every scene of the Tanks\&Temples dataset~\cite{tanksandtemples} in \cref{suptab:tandt}.

\begin{table*}
    \centering
    \resizebox{0.9\linewidth}{!}{
\setlength{\tabcolsep}{2pt}
\newdimen\wate \wate=30pt
\newdimen\wperc \wperc=22pt

\begin{tabular}{p{10pt}p{10pt}lcccccc@{\hskip 0.8em}|@{\hskip 0.8em}cccccc@{\hskip 0.8em}|@{\hskip 0.8em}cccccc@{\hskip 0.8em}|@{\hskip 0.8em}ccccccc}
\toprule
 &  &  & \multicolumn{6}{c}{\textbf{ATE ($\downarrow$)}} & \multicolumn{6}{c}{\textbf{RTA@5 ($\uparrow$)}} & \multicolumn{6}{c}{\textbf{RRA@5 ($\uparrow$)}} & \multicolumn{6}{c}{\textbf{Reg. ($\uparrow$)}} \\
 &  &  Scene & \rotatebox{90}{COLMAP~\cite{colmapsfm}} & \rotatebox{90}{ACE-Zero~\cite{acezero}} & \rotatebox{90}{FlowMap~\cite{flowmap}} & \rotatebox{90}{VGGSfM~\cite{vggsfm}} & \rotatebox{90}{DF-SfM~\cite{detfreeSfm}} & \rotatebox{90}{\textbf{MASt3R-SfM}} & \rotatebox{90}{COLMAP~\cite{colmapsfm}} & \rotatebox{90}{ACE-Zero~\cite{acezero}} & \rotatebox{90}{FlowMap~\cite{flowmap}} & \rotatebox{90}{VGGSfM~\cite{vggsfm}} & \rotatebox{90}{DF-SfM~\cite{detfreeSfm}} & \rotatebox{90}{\textbf{MASt3R-SfM}} & \rotatebox{90}{COLMAP~\cite{colmapsfm}} & \rotatebox{90}{ACE-Zero~\cite{acezero}} & \rotatebox{90}{FlowMap~\cite{flowmap}} & \rotatebox{90}{VGGSfM~\cite{vggsfm}} & \rotatebox{90}{DF-SfM~\cite{detfreeSfm}} & \rotatebox{90}{\textbf{MASt3R-SfM}} & \rotatebox{90}{COLMAP~\cite{colmapsfm}} & \rotatebox{90}{ACE-Zero~\cite{acezero}} & \rotatebox{90}{FlowMap~\cite{flowmap}} & \rotatebox{90}{VGGSfM~\cite{vggsfm}} & \rotatebox{90}{DF-SfM~\cite{detfreeSfm}} & \rotatebox{90}{\textbf{MASt3R-SfM}} \\
\bottomrule
\end{tabular}
}
\end{table*}
\begin{table*}
    \centering
    \resizebox{0.9\linewidth}{!}{
\setlength{\tabcolsep}{2pt}
\newdimen\wate \wate=30pt
\newdimen\wperc \wperc=22pt

\begin{tabular}{p{10pt}p{10pt}lcccccc@{\hskip 0.8em}|@{\hskip 0.8em}cccccc@{\hskip 0.8em}|@{\hskip 0.8em}cccccc@{\hskip 0.8em}|@{\hskip 0.8em}cccccc}
\toprule
 &  &  & \multicolumn{6}{c}{\textbf{ATE ($\downarrow$)}} & \multicolumn{6}{c}{\textbf{RTA@5 ($\uparrow$)}} & \multicolumn{6}{c}{\textbf{RRA@5 ($\uparrow$)}} & \multicolumn{6}{c}{\textbf{Reg. ($\uparrow$)}} \\
 &  &  Scene & \rotatebox{90}{COLMAP~\cite{colmapsfm}} & \rotatebox{90}{ACE-Zero~\cite{acezero}} & \rotatebox{90}{FlowMap~\cite{flowmap}} & \rotatebox{90}{VGGSfM~\cite{vggsfm}} & \rotatebox{90}{DF-SfM~\cite{detfreeSfm}} & \rotatebox{90}{\textbf{MASt3R-SfM}} & \rotatebox{90}{COLMAP~\cite{colmapsfm}} & \rotatebox{90}{ACE-Zero~\cite{acezero}} & \rotatebox{90}{FlowMap~\cite{flowmap}} & \rotatebox{90}{VGGSfM~\cite{vggsfm}} & \rotatebox{90}{DF-SfM~\cite{detfreeSfm}} & \rotatebox{90}{\textbf{MASt3R-SfM}} & \rotatebox{90}{COLMAP~\cite{colmapsfm}} & \rotatebox{90}{ACE-Zero~\cite{acezero}} & \rotatebox{90}{FlowMap~\cite{flowmap}} & \rotatebox{90}{VGGSfM~\cite{vggsfm}} & \rotatebox{90}{DF-SfM~\cite{detfreeSfm}} & \rotatebox{90}{\textbf{MASt3R-SfM}} & \rotatebox{90}{COLMAP~\cite{colmapsfm}} & \rotatebox{90}{ACE-Zero~\cite{acezero}} & \rotatebox{90}{FlowMap~\cite{flowmap}} & \rotatebox{90}{VGGSfM~\cite{vggsfm}} & \rotatebox{90}{DF-SfM~\cite{detfreeSfm}} & \rotatebox{90}{\textbf{MASt3R-SfM}} \\
\bottomrule
\end{tabular}
}
    \vspace{-0.25cm}
    \caption{\textbf{Detailed per-scene results on Tanks \& Temples} in terms of ATE, pose accuracy (RTA@5 and RRA@5) and registration rate (Reg.).
        For easier readability, we color-code the results as a linear gradient between 
        \BGcolor{e0a4a4}{ w}\BGcolor{e7b2ac}{o}\BGcolor{eec0b5}{r}\BGcolor{f6cebe}{s}\BGcolor{fcdcc6}{t}\BGcolor{ffe4ca}{ }\BGcolor{fee8cc}{a}\BGcolor{feeccd}{n}\BGcolor{fdf0cf}{d}\BGcolor{f6f0cb}{ }\BGcolor{e3e6bd}{b}\BGcolor{cfdcae}{e}\BGcolor{bbd2a0}{s}\BGcolor{a8c992}{t }
        per-row result for that metric. Reg. is color-coded with linear gradient between 
        \BGcolor{e0a4a4}{ 0}\BGcolor{e9b6af}{\%}\BGcolor{f3c9ba}{ }\BGcolor{fcdbc5}{a}\BGcolor{ffe5ca}{n}\BGcolor{feeacc}{d}\BGcolor{fdefcf}{ }\BGcolor{f4efc9}{1}\BGcolor{dbe2b7}{0}\BGcolor{c1d5a4}{0}\BGcolor{a8c992}{\%. } We mark missing results with \BGcolor{eeeeee}{ - } (not converged / runtime errors / ground truth).
        }
    \vspace{-0.25cm}
    \label{suptab:tandt}
\end{table*}

\section{Additional ablations}

\begin{figure}
    \centering
    \vspace{-0.15cm}
    \includegraphics[width=\linewidth]{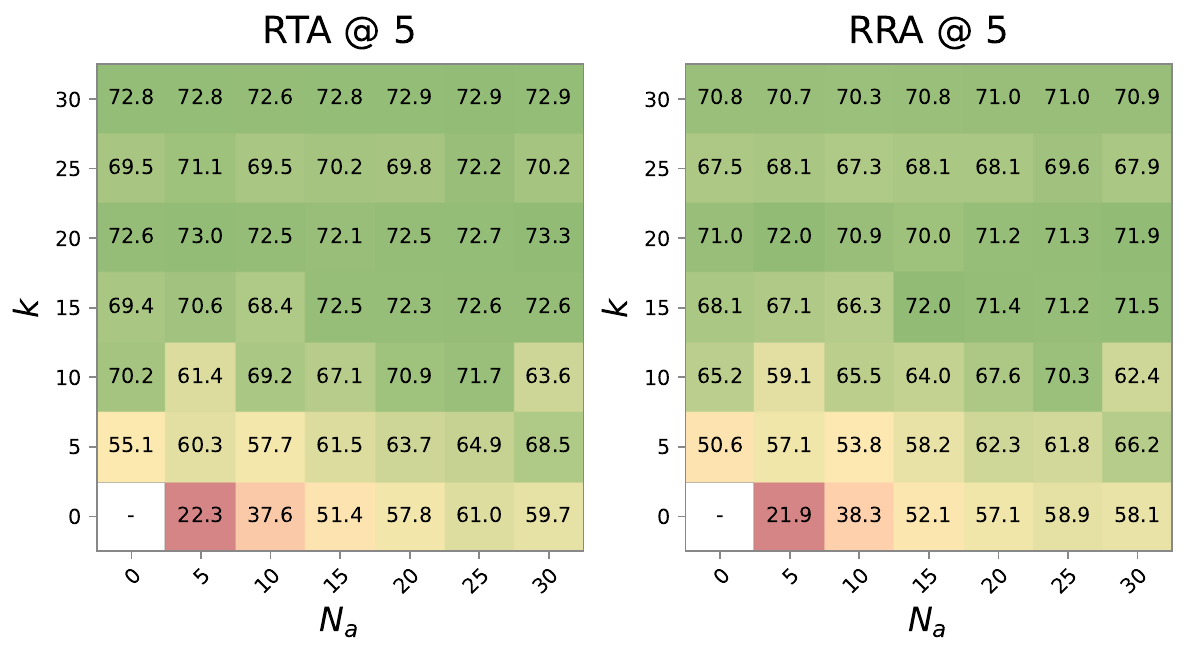}
    \vspace{-0.85cm}
    \caption{\textbf{Pose accuracy ($\uparrow$)} %
    on T\&T-200 \wrt the number of key images $N_a$ and number of nearest neighbors $k$}
    \vspace{-0.2cm}
    \label{fig:sup-abl-graph}
\end{figure}

We study the effect of varying the hyperparameters for the construction of the sparse scene graph (Section 4.1 of the main paper) in Fig.~\ref{fig:sup-abl-graph}. 
Generally increasing the number of key images ($N_a$) or nearest neighbors ($k$) leads to improvements in performance, which saturate above $N_a \ge 20$ or $k \ge 10$.

\section{Parametrizations of Cameras}

\begin{table}[t]
    \centering
    \resizebox{0.8\linewidth}{!}{\begin{tabular}{lccc}
\toprule
 & ATE$\downarrow$ & RTA@5$\uparrow$ & RRA@5$\uparrow$ \\
\midrule
\multicolumn{2}{@{}l}{Camera reparametrization} \\
No & 0.01445 & 56.0 & 52.5 \\
\textbf{Yes} & \textbf{0.01243} & \textbf{70.9} & \textbf{67.6} \\

\midrule
\multicolumn{2}{@{}l}{Kinematic chain} \\
No & 0.01675 & 52.2 & 50.0 \\
Star & 0.02013 & 42.0 & 39.2 \\
MST & 0.01600 & 64.4 & 62.1 \\
H. clust. (sim) & 0.01517 & 64.2 & 62.6 \\
\textbf{H. clust (\#corr)} & \textbf{0.01243} & \textbf{70.9} & \textbf{67.6} \\

\bottomrule
\end{tabular}
}
    \vspace{-0.25cm}
    \caption{Effects of camera reparametrization and kinematic chain on T\&T-200.}
    \label{tab:sup-abl-kinmode}
\end{table}

As noted by other authors~\cite{camp}, a clever parametrization of cameras can significantly  accelerate convergence.
In the main paper, we describe a camera $\mathcal{K}_n = (\K{n},\cam{n})$ classically as intrinsic and extrinsic parameters, where 
\begin{equation}
    \K{n} = \left[\begin{array}{ccc}
f_n & 0 & c_{x}\\
0 & f_n & c_{y}\\
0 & 0 & 1
\end{array}\right] \in \R^{3\times3},
\end{equation}
\begin{equation}
    \cam{n} = \left[\begin{array}{c|c}
R_n & t_n\\
 0  & 1
\end{array}\right] \in \R^{4\times4}.
\end{equation}
Here, $f_n>0$ denotes the camera focal, $(c_x,c_y)=(\nicefrac{W}{2},\nicefrac{H}{2})$ is the optical center, $R_n \in \R^{3\times3}$ is a rotation matrix typically represented as a quaternion $q_n\in\R^4$ internally, and $t_n\in\R^3$ is a translation.

\mypar{Camera parametrization.}

During optimization, 3D points are constructed using the inverse reprojection function $\pi^{-1}(\cdot)$ as a function of the camera intrinsics $\K{n}$, extrinsics $\cam{n}$, pixel coordinates and depthmaps $\depth{n}$ (see Section 4.2 from the main paper).
One potential issue with this classical parametrization is that small changes in the extrinsics can typically induce a large change in the reconstructed 3D points.
For instance, small noise on the rotation $R_n$ could result in a potentially large absolute motion of 3D points, motion whose amplitude would be proportional to the points' distance to camera (\ie their depth).
It seems therefore natural to reparametrize cameras so as to better balance the variations between camera parameters and 3D points.
To do so, we propose to switch the camera rotation center from the optical center to a point `in the middle' of the 3D point-cloud generated by this camera, or more precisely, at the intersection of the $\overrightarrow{z}$ vector from the camera center and the median depth plane.
In more details, we construct the extrinsics $\cam{n}$ using a fixed post-translation $\tilde{T}_n\in\R^4$ on the $z$-axis as as $\cam{n} \eqdef T_n \cam{n}'$, with
\begin{equation}
\tilde{T}_n = \left[\begin{array}{cccc}
1 & 0 & 0 & 0\\
0 & 1 & 0 & 0\\
0 & 0 & 1 & \tilde{m}_n^z\\
0 & 0 & 0 & 1\\
\end{array}\right],
\end{equation}
where $\tilde{m}_n^z=\text{median}(\cdepth{n}) f_n/\tilde{f}_n$ is the median canonical depth for image $\I{n}$ modulated by the ratio of the current focal length \wrt the canonical focal $\tilde{f}_n$, and $\cam{n}'$ is again parameterized as a quaternion and a translation.
This way, rotation and translation noise in $R_n$ are naturally compensated and have a lot less impact on the positions of the reconstructed 3D points, as illustrated in \cref{tab:sup-abl-kinmode}.

\mypar{Kinematic chain.}
A second source of undesirable correlations between camera parameters stems from the intricate relationship between overlapping viewpoints.
Indeed, if two views overlap, then modifying the position or rotation of one camera will most likely also result in a similar modification of the second camera, since the modification will impact the 3D points shared by both cameras.
Thus, instead of representing all cameras independently, we propose to express them relatively to each other using a kinematic chain.
This naturally conveys the idea than modifying one camera will impact the other cameras \emph{by design}.
In practice, we define a \emph{kinematic tree} $\T=(\V,\TE)$ over all cameras $\V$.
$\T$ consists of a single root node $r\in\V$ and a set of directed edges $(n \rightarrow m)\in\TE$, with $|\TE|=N-1$ since $\T$ is a tree.
The pose of all cameras is then computed in sequence, starting from the root as
\begin{equation}
    \forall (n \rightarrow m)\in\TE,~
    \cam{m} = \cam{n \rightarrow m} \cam{n}. %
\end{equation}
Internally, we thus only store as free variables the set of poses $\{\cam{r}\} \cap \{\cam{n\rightarrow m}\}_{(n \rightarrow m)\in\TE}$, each one represented as mentioned above.
In the end, this parametrization results in exactly the same number of parameters as the classical one.

We experiment with different strategies to construct the kinematic tree $\T$ and report the results in \cref{tab:sup-abl-kinmode}: `star' refers to a baseline where $N-1$ cameras are connected to the root camera, which performs even worse than a classical parametrization; `MST' denotes a kinematic tree defined as maximum spanning tree over the similarity matrix $S$; and `H. clust.' refers to a tree formed by hierarchical clustering using either raw similarities from image retrieval or actual number of correspondences after the pairwise forward with \master{}.
This latter strategy performs best and significantly improves over previous baselines, highlighting the importance of a balanced graph with approximately $\log_2(N)$ levels (in comparison, a star-tree has just 1 level, while a MST tree can potentially have $N/2$ levels at most).
Note that the sparse scene graph $\G$ from section 4.1 and the kinematic tree $\T$ share no relation other than being defined over the same set of nodes.

{
    \small
    \bibliographystyle{ieeenat_fullname}
    \bibliography{main}
}
\end{document}